\documentclass[letterpaper]{article}
\usepackage[preprint]{aaai2027}
\usepackage[hyphens]{url}
\usepackage{graphicx}
\usepackage{natbib}
\usepackage{caption}
\usepackage{amsmath}
\usepackage{amssymb}
\usepackage{booktabs}
\usepackage{multirow}
\usepackage{makecell}
\usepackage{microtype}
\usepackage{subcaption}
\usepackage[utf8]{inputenc}

\title{M-SQE: Multilingual Skill Quality Estimation for Enhancing Language Equality in Agentic Skill Use}

\author{
    Yilun Liu, Shimin Tao, Minggui He, Chenxin Liu, Li Zhang, Chen Liu,\\
    Miao Zhang, Jiaxin Guo, Min Zhang, Liqun Deng, Xiaojun Meng, Daimeng Wei
}

\affiliations{
    Huawei, China\\
    \texttt{liuyilun3@huawei.com}
}

\begin{document}

\maketitle

\begin{abstract}
Agent skills, reusable procedural documents that extend LLM agents beyond their parametric memory, have become an important interface for deploying agents on real-world tasks. Community-maintained skill libraries built around this interface are growing rapidly. However, this ecosystem remains deeply English-centric: our audit finds that low-resource languages such as Swahili and Hindi have no in-language skill content, so retrieval often returns a skill written in a different language than the query, degrading accuracy and recall. A practical solution is to synthesize in-language skills for retrieval but the quality can be unreliable, so relevance in this setting alone often surfaces a related but unusable candidate. To address this, we propose M-SQE, a post-retrieval Multilingual Skill Quality Estimation framework that scores candidates via a Theory view for intrinsic quality and an Action view for task-grounded utility, unified into a domain-conditioned final score. We evaluate M-SQE across three skill-use domains: general, tool-use, and cultural tasks. Empirically, we build three-layer candidate skill pools mirroring today's ecosystem, where M-SQE's task success exceeds existing baseline's average by at least +3.5 points across three different retrievers. Particularly, M-SQE lifts the lowest-resource languages most (+12.9pp on Hindi and +5.6pp on Swahili) and achieves strong performance across all six culture regions, thereby moving agentic skill use toward linguistic and cultural equality.

\end{abstract}

\section{Introduction}
\label{sec:intro}

Agent skills are reusable procedural documents that package step-by-step instructions, function schemas, API conventions, and domain knowledge. They have become an important interface for extending large language model (LLM) agents beyond their parametric memory, letting them act reliably on multi-step tasks \citep{wang2023voyager,xu2026agentskills,zhou2026agentskillssurvey}. Community-maintained skill libraries built around this interface have grown quickly, but almost entirely in English. Table~\ref{tab:audit} reports our audit spanning over a dozen public community skill indexes ($\sim$84,700 entries in total, including a popular 38.2k-star skill repository) \citep{littledinoc2026agentskills,wshobson2026agents,huzey2026claudeskills}: genuine target-language skill content is concentrated in a handful of high-resource languages, while Swahili and Hindi, spoken natively by hundreds of millions of people combined, have zero target-language skill content of their own (full audit protocol in Appendix~\ref{sec:app-audit}). In practice, an agent serving these languages has no in-language skill to retrieve.

\begin{table}[t]
\centering\small
\begin{tabular}{lr}
\toprule
Language & Est.\ in-language skill bodies \\
\midrule
English (en) & $\sim$79,700 \\
\midrule
Chinese (zh) & $\sim$4,000 \\
French (fr) & $\sim$700 \\
Korean (ko) & $\sim$200 \\
Japanese (ja) & $\sim$80 \\
Swahili (sw) & 0 \\
Hindi (hi) & 0 \\
\bottomrule
\end{tabular}
\caption{Estimated genuine in-language skill bodies among $\sim$84,700 audited community skill entries: \textbf{the ecosystem is mostly English, while Swahili and Hindi have zero.}}
\label{tab:audit}
\end{table}

\begin{figure}[t!]
\centering
\includegraphics[width=0.75\linewidth]{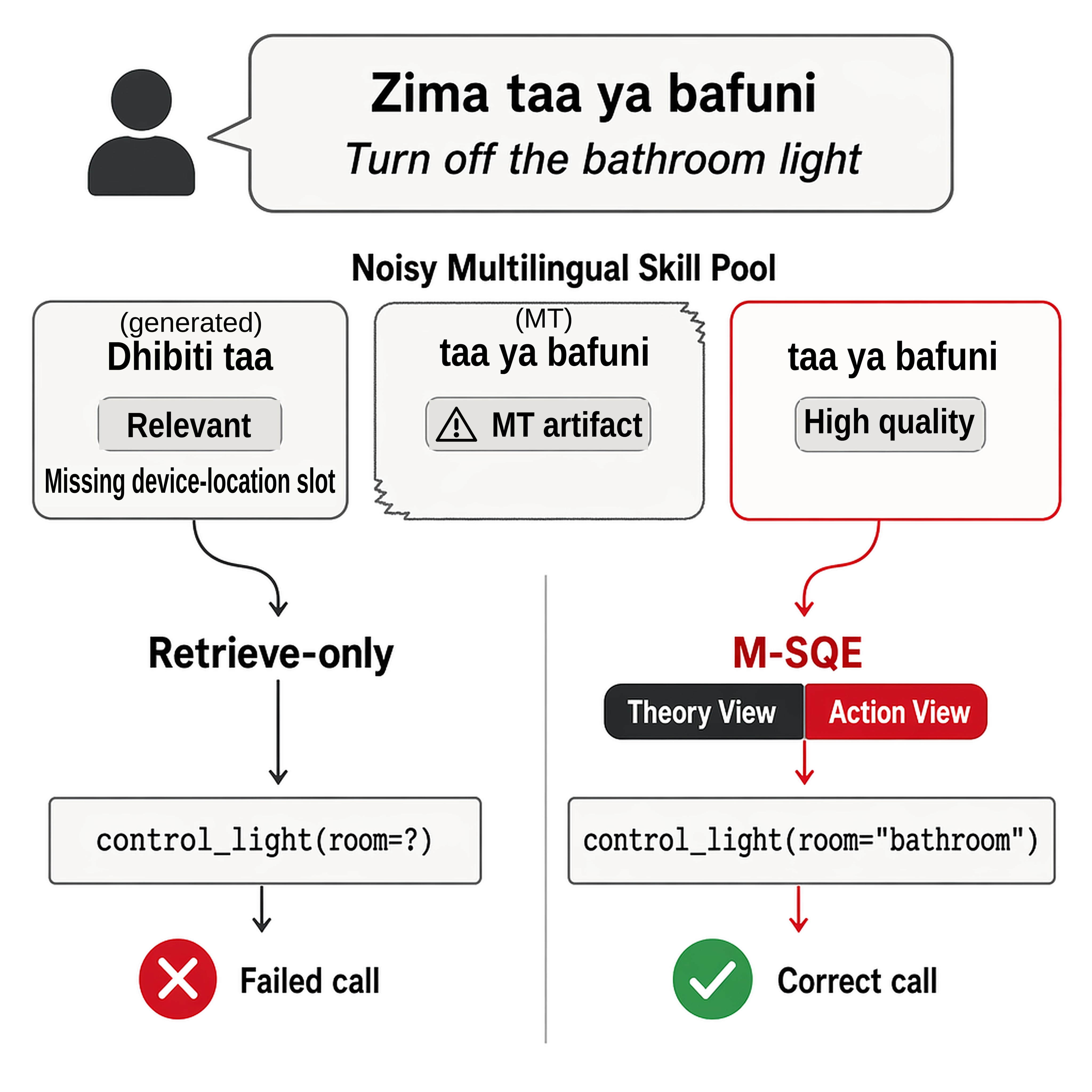}
\caption{A representative failure for a Swahili speaker: retrieve-only selection surfaces a topically relevant smart-home skill that omits a slot required for execution. \textbf{M-SQE's Theory view screens out this gap and its Action view ranks the rest by fit, selecting a skill the agent can execute.}
}
\label{fig:banner}
\end{figure}

However, an English-only skill library is not merely an inconvenience: \textbf{it actively degrades agent performance} for non-English users. \citet{xbcp2026} show that when a query and its supporting evidence are written in different languages, both answer accuracy and evidence recall drop sharply, possibly due to the distribution mismatch introduced by cross-lingual retrieval.
 Skills play exactly this evidentiary role for an agent. An agent that retrieves an English
 skill for a Swahili or Hindi query therefore inherits the same mismatch failure mode, owing to the absence of native-language skills.
 This is, in effect, an agent-era language equality problem: languages already underserved by the web are now underserved by the very tools meant to make agents useful to their speakers \citep{joshi2020state,stanford2025digitaldivide}. Fig.~\ref{fig:banner} illustrates the resulting failure mode and how our proposed M-SQE avoids it.

A practical way to close this gap is to synthesize more multilingual skills \citep{long2024llmsyndata,ma2026skillgen,wang2026skillx}, but doing so exposes two further challenges in an actual agentic skill-use scenario, where an agent has to choose the most suitable skill from a large pool. \textbf{The first is quality:}
 multilingual skills are typically synthesized by machine translation (MT) or model self-generation, and neither is stable in quality. \citet{skillsbench2026} find that human-curated skills help agents
 on average while self-generated skills can hurt task performance due to unstable quality. \citet{liu2026midb} document that MT content can introduce content errors, wrong-language artifacts, and insufficient cultural localization. \textbf{The second challenge} is that relevance does not imply usability. Even in English-only settings, Li et al.~(\citeyear{skillflow2025}) report that agents adopt a relevance-retrieved skill 70.1\% of the time with no accompanying performance gain. Realistic multilingual pools an agent might face, which mix ecological-style, MT, and self-generated material, widen this gap further: with lower multilingual quality, even more of these skills look relevant but provide no benefit.

To address this, we introduce M-SQE, a post-retrieval Multilingual Skill Quality Estimation framework that scores every retrieved candidate from two complementary views. A Theory view estimates a skill's intrinsic, task-independent quality, and an Action view estimates its grounded utility for the query at hand. Because what makes a skill usable varies by task (\emph{e.g.}, tool use hinges on a strict execution contract, while cultural queries hinge on corroborated evidence), a router conditions how the two views combine into the final score, closing the gap between what retrieval finds and what an agent can actually execute. Our contributions are:

\begin{itemize}
    \item We propose M-SQE, to our knowledge the first post-retrieval quality estimation framework purpose-built for multilingual agent skills, filling the usability gap of retrieve-only approaches with an average +6.3pp task-success gain across all nine evaluation settings.
    \item We demonstrate that M-SQE narrows the language inequality of agentic skill use with gains on lowest-resource languages (+12.9pp/+5.6pp task success on Hindi/Swahili).
    \item We open-source our evaluation set, skill pools, and M-SQE implementation, facilitating future research on multilingual and cross-cultural agent skill use\footnote{Available at \url{https://github.com/lunyiliu/M-SQE}.}.
\end{itemize}

\section{Social Impact of M-SQE}
\label{sec:impact}

M-SQE targets a concrete instance of linguistic and cultural inequality in agentic AI: the skills that let language agents act reliably are overwhelmingly written in English, leaving speakers of low-resource languages and members of non-Western cultural communities with agents that are, in practice, less capable on their behalf. By aiding agentic skill retrieval in a noisy, realistic multilingual pool through two-view quality estimation, M-SQE advances agent-era equality in two ways:

(1) \textbf{Bridging the Skill Divide for Underserved Language Communities.} Non-English speakers already face a documented digital divide at the model layer \citep{stanford2025digitaldivide}; as Section~\ref{sec:intro}'s audit finds, that divide recurs one layer further down the agent stack, where Swahili and Hindi have zero genuine in-language skill content. Speakers of these languages are therefore excluded from the agent-era productivity dividend not by the model they talk to, but by the skill library the agent draws on to act on their behalf. M-SQE addresses this gap effectively. On agentic tool-use tasks, M-SQE lifts Hindi and Swahili most (Fig.~\ref{fig:equity}): task success rises by +12.9pp on Hindi and +5.6pp on Swahili. A skill pool that was only usable by chance becomes one an agent serving an underserved-language community can actually rely on.

(2) \textbf{Advancing Cultural Equality in Agentic AI.} Multilingual capability is not the same as cultural competence \citep{rystrom2025multilingual}: an MT or self-generated skill can preserve fluent language while omitting or mis-attributing the cultural detail that determines whether an answer is correct. The Lunar New Year red-envelope case in Section~\ref{sec:method} illustrates the failure: a topically matched skill can still specify the wrong customary amount. M-SQE's cultural gains hold with stable deltas across six main culture regions, matching or exceeding both baselines in every region, with strong gains in South Asia (+6.6pp) and Oceania (+8.4pp). This region-by-region consistency, not a single number, is the evidence that M-SQE's quality signal captures cultural correctness rather than surface fluency.

\section{Related Work}
\label{sec:related}

\paragraph{Skill-Based Agents and Skill Retrieval.} Reusable procedural knowledge was popularized for embodied LLM agents by \citet{wang2023voyager}, whose skill library lets an agent accumulate and reuse verified action sequences during exploration. This idea has since grown into a community-scale ecosystem of packaged, on-demand agent skills \citep{xu2026agentskills}, and \citet{skillsbench2026} show that curated skills raise agent task success while self-generated ones offer little or even negative benefit, underscoring that not every skill in a growing pool is worth using. Li et al.~(\citeyear{skillflow2025}) focus on scaling skill \emph{retrieval} itself, with a four-stage, skill-specific retrieval pipeline that ranks candidates for the query at hand. Generic rerankers sharpen the retrieved list along the same axis: ToolRerank \citep{zheng2024toolrerank} adapts hierarchy-aware reranking to tool retrieval, while mMARCO \citep{bonifacio2021mmarco} extends relevance reranking across languages.

\paragraph{Multilingual Data Quality.} A parallel line of work has documented that scaling multilingual instruction data by MT or model self-generation, the two pipelines most multilingual synthesis efforts rely on, degrades quality rather than merely diluting fluency. \citet{lai2024llms} report substantial per-language MT error rates, and \citet{liu2026midb} and \citet{zhao2026mdaq} build on this observation with expert-revision and quality-scoring pipelines that clean multilingual instruction corpora before instruction tuning; quality scorers such as DEITA \citep{liu2024deita} likewise select instruction data on intrinsic quality alone. These pipelines judge only the data's own language quality, leaving out skill-use dimensions such as executability and context efficiency.

\paragraph{Positioning M-SQE.} The two lines above leave complementary gaps. The skill-retrieval line ranks candidates by how well they match the query, and relevance is where its judgment stops: it asks neither whether a matched skill is intrinsically sound, nor whether the skill will actually work for the task at hand. The data-quality line, in turn, stops at the data's own language quality. M-SQE fills both gaps: the Theory view brings intrinsic quality estimation to skill-use time, and the Action view adds what neither line measures --- how much a specific skill will help the query at hand, together with an explicit misleading-risk estimate.

\section{The M-SQE Framework}
\label{sec:method}

\begin{figure}[t]
\centering
\includegraphics[width=0.9\linewidth]{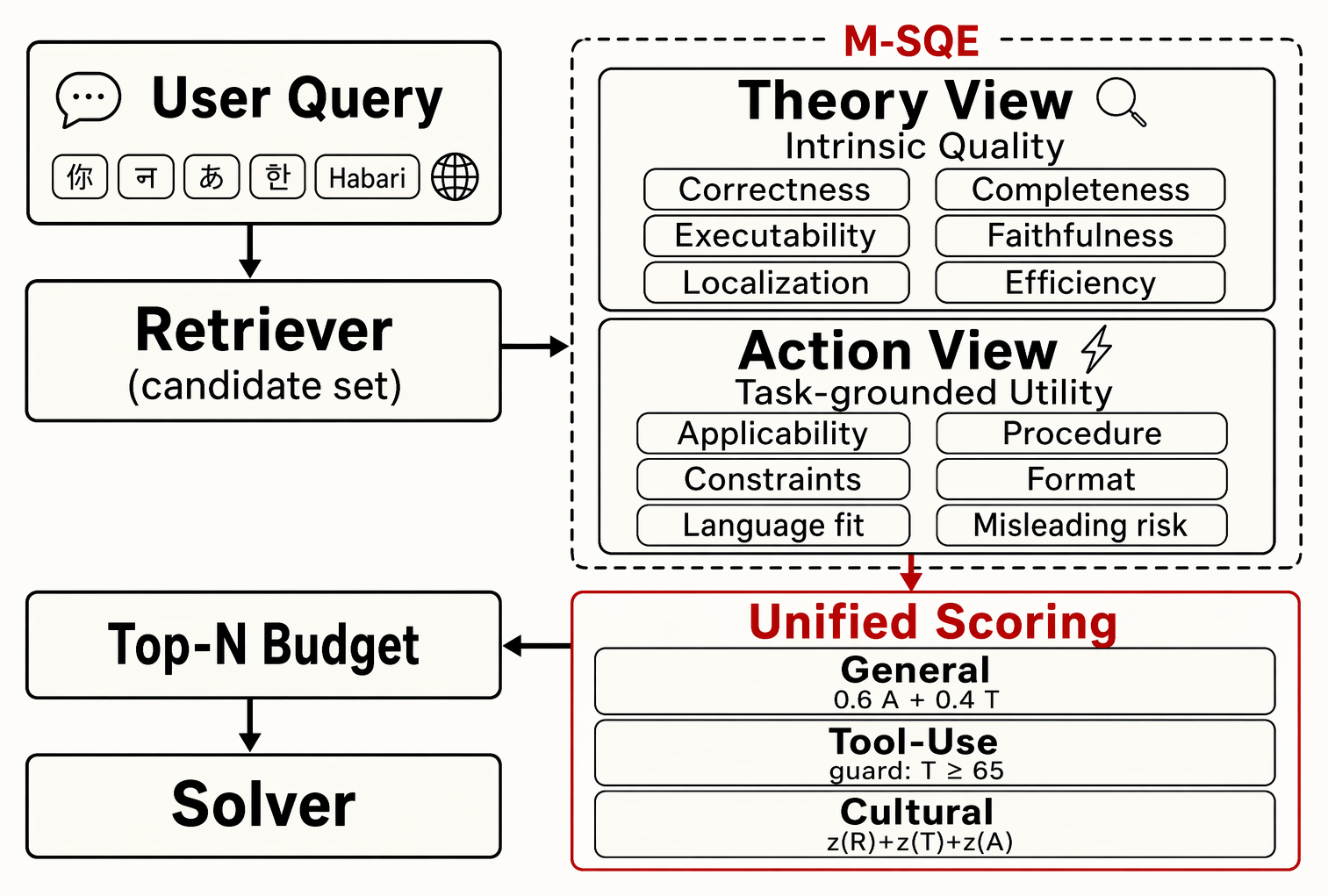}
\caption{Overview of M-SQE: retrieved skill candidates are scored by Theory and Action views, fused by the domain-routed unified score, and delivered to the solver for tasks.}
\label{fig:framework}
\end{figure}

\paragraph{Overview.} Fig.~\ref{fig:framework} gives an overview of M-SQE. Given a query, a retriever first returns a candidate set of skills. M-SQE then scores every candidate from two complementary views: a \textit{Theory view} that certifies intrinsic skill quality and an \textit{Action view} that estimates task-grounded utility. The two views are routed into a single score based on the task domain at hand, and the resulting ranking determines which skills, within a fixed Top-$N$ budget, are handed to the downstream solver (the agent that executes the user's task).

\subsection{Problem Formulation}
Given a task query $q$, a skill pool $S$, a retriever $R$, and a candidate depth $K$, the retriever returns a candidate set $C_K = R(q,S,K)$. M-SQE then selects a fixed-size subset $Z_N \subset C_K$, with budget $N \le K$, and the downstream solver receives only the query and the anonymized skill bodies in $Z_N$ (provenance labels and source identifiers stripped), without any gold answer or checker metadata. The objective is to maximize deterministic task success under the same skill-body budget. Retrieval decides which candidates are visible; M-SQE decides which visible candidates are usable.

\subsection{Theory View}
The Theory view certifies intrinsic skill quality independent of any particular query, scoring six dimensions:

(1) \textit{Correctness} (red-line): the skill's procedural facts and logic are right; (2) \textit{Completeness}: the necessary steps and boundary cases of the task class are covered; (3) \textit{Executability}: steps are concrete enough for an agent to act on, not merely descriptive; (4) \textit{Cross-lingual faithfulness}: no meaning-altering mistranslation or untranslated fragments; (5) \textit{Localization}: the prose is genuinely written for the target language; (6) \textit{Context efficiency}: concise enough not to dilute agent's attention.

These dimensions are inspired by existing multilingual instruction-data quality taxonomy~\citep{liu2026midb}, and are further extended to the demands of multilingual agent skills. A language-service expert team averaging over six years of professional experience constructs the dimensions across multiple rounds of review, organizing them into two blocks: general quality dimensions --- correctness, completeness, cross-lingual faithfulness, and localization --- and agent-specific dimensions --- executability and context efficiency in Theory view (as well as the six dimensions in Action view)--- that instruction data does not require but an executing agent does. The overall Theory score is the six-dimension mean on a 0--100 scale, capped at 40 if the red-line dimension is violated and at 80 if any basic dimension is, so a single severe defect cannot be averaged away. Full expert profiles, the expert construction process behind both Thoery and Action views' dimensions, per-dimension cases, and the full scoring rubric are in Appendix~\ref{sec:app-prompts}.

\subsection{Action View}
The Action view estimates task-grounded utility for the current query and a candidate skill, scoring five grounded-fit dimensions plus an explicit risk estimate:

(1) \textit{Task applicability}: whether the skill targets this task at all; (2) \textit{Procedure match}: the steps cover what the query actually requires; (3) \textit{Constraint match}: slots, parameters, and constraints line up with the request; (4) \textit{Output-format match}: the skill produces the output shape the task expects; (5) \textit{Language fit}: the prose is usable in the query's language; (6) \textit{Misleading risk}: the chance that a related-looking skill steers the solver to a wrong API, value, or convention.

These dimensions targets the adoption gap documented by Li et al.~(\citeyear{skillflow2025}), where skills look relevant enough to be adopted yet fail to help. Each dimension is scored 0–100. Consequently, the rubric permits a high Action score only when the procedure is executable, the constraints match, and the misleading risk is low, avoiding trapped by surface relevance.

\subsection{Unified Scoring}
Based on the expert-examined dimensions above, we prompt a scorer LLM per view with its full rubric, obtaining the Theory and Action scores used throughout this section. M-SQE evaluates every candidate skill $s$ under the same Top-$N$ budget with a unified score that routes three domain-specific scoring rules, $C_{\text{gen}}$, $C_{\text{func}}$, and $C_{\text{cul}}$, through task domain $\tau$:
\begin{equation}
S_{\text{M-SQE}}(q,s;\tau) = a_\tau C_{\text{gen}}(q,s) + b_\tau C_{\text{func}}(q,s) + c_\tau C_{\text{cul}}(q,s),
\label{eq:unified}
\end{equation}
where $(a_\tau,b_\tau,c_\tau)\in\{0,1\}^3$ with $a_\tau+b_\tau+c_\tau=1$. The indicator tuple is fixed directly by the task domain $\tau$: the general domain sets $(1,0,0)$, the tool-use domain sets $(0,1,0)$, and the cultural domain sets $(0,0,1)$. Each of the three components combines Theory and Action (and, for cultural tasks, Retrieval) in a way matched to the nature of the task.

\paragraph{Domain Router.} Multilingual Agentic skill use can be roughly divided into three classes: tool-heavy tasks and culture-heavy tasks, each has its own focus, and a general class covering the rest. Based on this taxonomy, the task domain $\tau$ is predicted from the query by a 5-shot prompted LLM domain router (five examples per domain written from the public task definitions alone), reaching 98.5\% accuracy on our three-domain query set. A noise analysis shows the router's labels can be randomly flipped on up to 90\% of queries while M-SQE stays ahead of Retrieve-only throughout (Fig.~\ref{fig:router}). A new specialized domain can be added the same way: a new adapter paired with five added examples.

\paragraph{General-task convex fusion.} $C_{\text{gen}} = 0.6\,\text{Action} + 0.4\,\text{Theory}$. General skill-use tasks cover heterogeneous reusable procedures (spreadsheet formulas, media-processing recipes, code patterns) whose usefulness is graded rather than binary: a skill can be comprehensive yet miss the one operator a query needs, or read as directly relevant while a small execution detail is wrong. The single mixing coefficient gives task applicability a modest majority while retaining intrinsic quality as a regularizer. For example, when a query asks for an \texttt{XLOOKUP} formula, a comprehensive spreadsheet overview that never states lookup semantics should not outrank a short, correct skill; conversely, a relevant-looking snippet with a malformed argument order is discounted by its Theory score.

\paragraph{Tool-use quality guarding.} Tool use carries a discrete execution contract: the selected skill must expose the correct function schema, required slots, and admissible value formats, so usefulness here is gated, not graded. $C_{\text{func}}$ ranks candidates by the lexicographic order $(\text{Theory}\geq 65,\ \text{Action})$, equivalently $C_{\text{func}} = M \cdot \mathbf{1}[\text{Theory} \geq 65] + \text{Action}$ for any $M$ larger than the Action range as a constant offset lifting every guard-passing candidate above all others. The threshold of 65 marks the boundary below which a skill is structurally unusable on the 0--100 Theory scale — for example, missing a required slot or carrying a critical localization failure — and candidates whose Theory scoring flags a language red-line are guarded out the same way. Consider a smart-home skill retrieved for \textit{``turn off the bathroom light''}: it may read as topically on point yet omit the device-location slot, leaving it unusable regardless of how relevant it looks. The guard removes such a candidate from contention first; Action then ranks the remaining executable skills by how well their slots match the request.

\paragraph{Cultural evidence fusion.} $C_{\text{cul}} = z(\text{Retrieval}) + z(\text{Theory}) + z(\text{Action})$, where $z(\cdot)$ denotes per-query z-score standardization within the retriever's candidate set. Culture questions fail in three independent ways — the retrieved skill can name the wrong event, state an unreliable fact, or fail to resolve the question actually asked — so no single view is sufficient and none should dominate the others' scale. A query about Lunar New Year red envelopes illustrates two of the three: a skill about wedding-gift etiquette can be factually sound yet grounded in the wrong occasion, while a self-generated Lunar New Year skill can name the right occasion and still misstate the customary amount. With equal standardized fusing, Retrieval anchors event grounding, Theory checks factual reliability, and Action checks whether the skill answers the question at hand.

Each component's own constants — the General fusion's 0.6/0.4 weight, the Tool-Use guard's Theory$\geq$65 threshold, and the Cultural fusion's equal standardized weights — follow from the design rationale above and stay stable under parameter perturbation (Appendix~\ref{sec:app-stability}).

\begin{table*}[t]
\centering\small
\begin{tabular}{l r p{3.6cm} p{5.2cm} r r}
\toprule
Domain & \#Tasks & Query language/culture & Three-layer skill pool & Pool size & Candidate depth \\
\midrule
General Skill Use & 94 & \raggedright fr, hi, ja, ko, sw, zh & \raggedright ecological-style 400 + MT 550 + self-generated 800 & 1,750 & 10 \\
Tool Use & 265 & \raggedright en, fr, hi, ja, ko, sw, zh & \raggedright ecological-style 1,254 + MT 660 + self-generated 385 & 2,299 & 20 \\
Cultural Skill Use & 52 & \raggedright Six main cultural regions & \raggedright ecological-style 3,384 + MT 1,283 + self-generated 618 & 5,285 & 50 \\
\bottomrule
\end{tabular}
\caption{Evaluation setup by domain. Each test task draws skills from a source-included three-layer skill pool (ecological-style, MT, and model self-generated); candidate depth is the retriever output size available to selectors.}
\label{tab:arms}
\end{table*}

\section{Experiments}
\label{sec:exp}

\subsection{Construction of Evaluation Tasks}

Our evaluation spans three high-frequency scenarios of agentic skill use: general procedural knowledge work, tool invocation, and culturally grounded interaction, all common surfaces for multilingual agents. The statistics of evaluation tasks as well as the paired skill pool are shown in Table~\ref{tab:arms}.

All three domains are built by one construction process. Source tasks are drawn from real, published benchmarks and culture resources (detailed below). To ensure every task tests skill use rather than the solver's parametric memory, we follow the checker-verified construction of SkillsBench \citep{skillsbench2026} and pass each source task through a three-stage filter: (1) verifiability, keeping only tasks whose success a deterministic checker can decide; (2) skill necessity, screening out tasks that saturate without any skill, which leaves the Prompt-only anchor well below ceiling in all three domains (General 53.2\%, Tool-Use 20.4\%, and Cultural 46.2\%; Table~\ref{tab:main}); and (3) leakage control, rewriting task prompts into natural user queries that copy no text from any pool skill and share no content-bearing terms with the answer key (Appendix~\ref{sec:app-pool}). Each domain then stratifies the admitted tasks across its languages, regions, and originating benchmarks. All task query rewriting and composition were carried out by the same multilingual expert team introduced in Section~\ref{sec:method}, through the same multiple rounds of review.

\paragraph{General Skill Use.} General Skill Use covers 94 curated multilingual skill-use tasks across six languages (\textit{fr}, \textit{hi}, \textit{ja}, \textit{ko}, \textit{sw}, \textit{zh}), built in the SkillsBench paradigm \citep{skillsbench2026}. Task types span spreadsheet formulas, code snippets, and document workflows, so the needed skill is reusable how-to procedural documentation, not a single fixed API call.

\paragraph{Tool Use.} Tool Use draws on real utterances from \citet{kulkarni2025massiveagents}'s 52-language function-calling benchmark, restricted to seven languages (\textit{en}, \textit{fr}, \textit{hi}, \textit{ja}, \textit{ko}, \textit{sw}, \textit{zh}) spanning a 55-function inventory across smart-home and IoT control, calendars and alarms, media playback, and everyday information queries. In tool-use tasks the agent must invoke device and service functions with exact schemas and slot values; the skills used in this evaluation are the documents that teach those invocations. Success on each utterance is graded by a deterministic function/slot checker, reflecting whether the agent executed the correct call.

\paragraph{Cultural Skill Use.} Cultural Skill Use comprises 52 short-answer tasks constructed from published culture resources, including NormAd~\citep{rao2024normad}, CultureBank~\citep{shi2024culturebank}, CulturALL~\citep{lin2026culturall}, CultureScope~\citep{zhang2025culturescope}, CultureAtlas~\citep{fung2024cultureatlas}, SAGE~\citep{guo2026sage}. The tasks span six cultural regions worldwide (East and Southeast Asia, South Asia, Europe, Africa and the Middle East, the Americas, and Oceania). Task
 types span etiquette, gift-giving, and dining customs alongside grounded regional facts, so the needed skill is culture-point documentation, not a generic overview. Each task is an independent, natural short-answer prompt derived from a culture point.

\subsection{Skill Pool Construction}

We build each domain's paired skill pool the way a multilingual pool is assembled in practice. The audit in Section~\ref{sec:intro} shows that genuine in-language skill content is scarce outside a handful of high-resource languages, so multilingual coverage typically comes from layering the two scalable production paths, MT and model self-generation, on top of the available ecological material. So our pools in Table~\ref{tab:arms} are three-fold:

(1) The \textbf{ecological-style layer} approximates the naturally available material an agent would encounter today; it contains document-derived skills, rendered from existing community skill files, official product skills, and API documentation, and background-derived skills distilled from domain background documentation \citep{xu2026agentskills}, preserving the coverage gaps and stylistic variation of real skill authorship. (2) \textbf{MT} is the first scalable path once target-language originals run out, and it might come with defects such as translationese~\citep{lai2024llms,liu2026midb}. Our MT layer preserves these naturally occurring cross-lingual transfer artifacts; concretely, MT skills are translated from each domain's English source material into its non-English target languages, so a domain's MT volume follows its language roster and available source material. (3) \textbf{Self-generation} is the second scalable path, and its limits are similarly documented --- unstable downstream gain due to occasional hallucinations \citep{skillsbench2026,zhang2026coevoskills}. Our self-generation layer preserves the naturally occurring incompleteness and execution errors this production path is known to produce; concretely, self-generated skills are written by an LLM prompted with the domain's public task specifications (\emph{e.g.}, task descriptions, function schemas, or culture-point summaries), revealing no answer and checker.

The exact composition differs by domain (Table~\ref{tab:arms}), reflecting the materials naturally available in each. The pool is also source-included: a skill relevant to a given query may already sit inside it, but every candidate is judged on its own content alone. Source inclusion is the standard convention in retrieval evaluation. The relevant document stays inside the searchable corpus rather than being held out, and skill-retrieval work follows the same convention: Li et al.~(\citeyear{skillflow2025}) inject their own oracle skills into the retrieval index in evaluation. Detailed skill construction and leakage control are in Appendix~\ref{sec:app-pool}.

\subsection{Retrievers and Baselines}

Because a quality-estimation layer must work regardless of which retriever supplies candidates, we evaluate M-SQE across three retrievers: BM25, the classic lexical retriever \citep{robertson2009bm25}; Neural, a hybrid dense retriever~\citep{luan2021sparse}; and SkillFlow, the state-of-the-art skill-specific retrieval pipeline (Li et al.\ \citeyear{skillflow2025}), faithfully reimplemented from its four-stage design. Together the three retrievers span lexical, neural, and skill-specific retrieval. Retriever candidate depth follows $K = 10 \cdot \lfloor |S| / 1000 \rfloor$, where $|S|$ is the domain's pool size. This rule trades off two failure modes: too shallow confounds selection quality with retrieval recall, since the relevant skill may never surface, while too deep inflates scoring cost and floods the candidate set with skills no ranking could rescue. All post-retrieval baselines (\textit{selectors}) score this same fixed candidate set of depth $K$ and output a final selection of Top $N$ skills.

\begin{figure*}[t]
\centering
\includegraphics[width=0.9\linewidth]{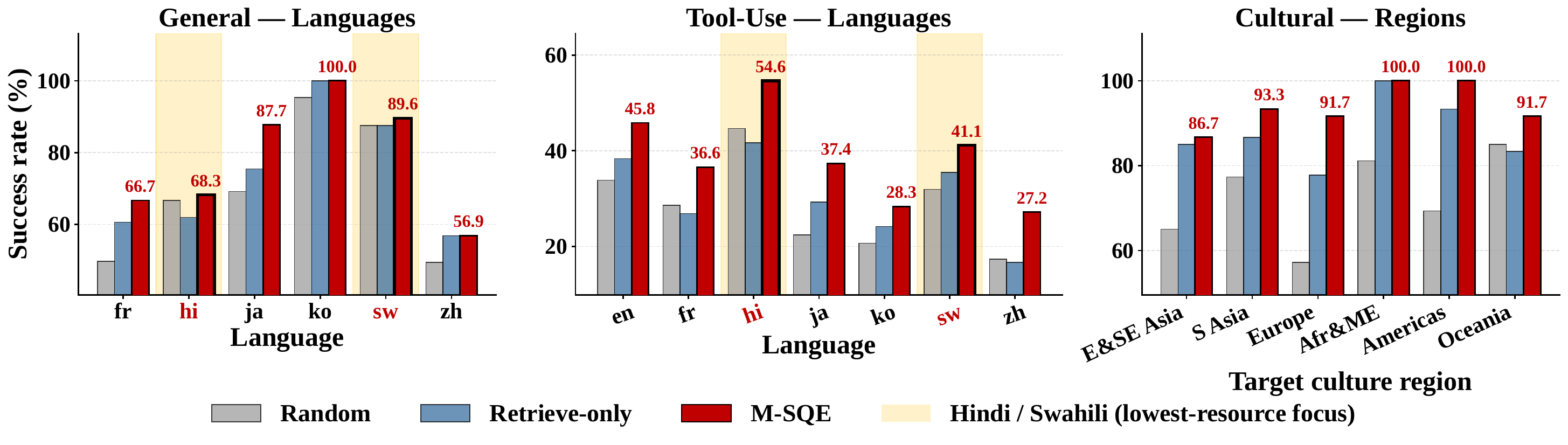}
\caption{Success rate across culture regions and languages: M-SQE matches or exceeds both baselines in every breakdown, with the largest gains on Hindi and Swahili, the two lowest-resource languages in our evaluation.}
\label{fig:equity}
\end{figure*}

For baselines compared with M-SQE, we reports two selection anchors, Random (a uniform selection from the candidate set) and Retrieve-only (the retriever's own ranking), together with six strong external baselines split into two families that both address important aspects of skill use. \textbf{Relevance rerankers} rank candidates by query-skill match: ToolRerank \citep{zheng2024toolrerank}, an adaptive, hierarchy-aware algorithmic reranker for tool retrieval; SkillFlow (Li et al.\ \citeyear{skillflow2025})'s own ranking stage, reused here as a post-retrieval selector regardless of which retriever supplied its candidates; and mMARCO \citep{bonifacio2021mmarco}, a multilingual cross-encoder reranker. \textbf{Quality scorers} assess intrinsic data quality of skills: multilingual data-quality scorers DEITA \citep{liu2024deita}, M-DaQ \citep{zhao2026mdaq}, and JQL \citep{ali2025jql}.

\begin{table}[t]
\centering\footnotesize
\resizebox{\linewidth}{!}{
\begin{tabular}{l@{\hskip 0.07in}c@{\hskip 0.07in}c@{\hskip 0.07in}c@{\hskip 0.07in}c@{\hskip 0.07in}c@{\hskip 0.07in}c@{\hskip 0.07in}c@{\hskip 0.07in}c@{\hskip 0.07in}c@{\hskip 0.07in}c}
\toprule
 & \multicolumn{3}{c}{General} & \multicolumn{3}{c}{Tool-Use} & \multicolumn{3}{c}{Cultural} & \multirow{2}{*}{Avg} \\
\cmidrule(lr){2-4} \cmidrule(lr){5-7} \cmidrule(lr){8-10}
Method & BM & Ne & SF & BM & Ne & SF & BM & Ne & SF & \\
\midrule
\multicolumn{11}{l}{\textit{Selection anchors}} \\
\midrule
Prompt-only & 53.2 & 53.2 & 53.2 & 20.4 & 20.4 & 20.4 & 46.2 & 46.2 & 46.2 & 39.9 \\
Random & 66.2 & 69.1 & 70.6 & 26.1 & 26.6 & 33.1 & 60.0 & 65.4 & 79.2 & 55.1 \\
Retrieve-only & \underline{71.3} & 70.2 & \underline{74.5} & 28.7 & 25.7 & \underline{37.4} & 78.8 & \underline{88.5} & 90.4 & 62.8 \\
\midrule
\multicolumn{11}{l}{\textit{Relevance rerankers}} \\
\midrule
ToolRerank & 64.9 & 73.4 & 70.2 & 30.6 & 26.0 & 34.7 & 76.9 & 86.5 & \underline{92.3} & 61.7 \\
SkillFlow & \underline{71.3} & \underline{74.5} & \underline{74.5} & \underline{33.2} & \textbf{36.2} & \underline{37.4} & \underline{86.5} & 86.5 & 90.4 & \underline{65.6} \\
mMARCO & 67.0 & 71.3 & 70.2 & 29.4 & 27.5 & 32.5 & \underline{86.5} & \textbf{90.4} & \underline{92.3} & 63.0 \\
\midrule
\multicolumn{11}{l}{\textit{Quality scorers}} \\
\midrule
DEITA & 68.1 & 61.7 & 69.1 & 30.2 & 29.1 & 35.1 & 51.9 & 51.9 & 75.0 & 52.5 \\
M-DaQ & 66.0 & 59.6 & 61.7 & 30.6 & \underline{31.7} & 34.3 & 63.5 & 57.7 & 76.9 & 53.5 \\
JQL & 59.6 & 69.1 & 70.2 & 25.3 & 20.4 & 33.2 & 53.8 & 57.7 & 75.0 & 51.6 \\
\midrule
\multicolumn{11}{l}{\textit{M-SQE (ours)}} \\
\midrule
\textbf{M-SQE} & \textbf{74.5} & \textbf{79.8} & \textbf{76.6} & \textbf{39.6} & \textbf{36.2} & \textbf{40.4} & \textbf{90.4} & \textbf{90.4} & \textbf{94.2} & \textbf{69.1} \\
\bottomrule
\end{tabular}
}
\caption{Task success rate (\%) at Top3, comparing M-SQE with existing skill-selection methods across three domains and three retrievers (BM=BM25, Ne=Neural, SF=SkillFlow). \textbf{Bold} = best in each column, \underline{underline} = second-best.}
\label{tab:main}
\end{table}

\subsection{Metrics and Backbones}

All metrics are deterministic checkers: task-specific checkers for General Skill Use and Cultural Skill Use, and an exact function/slot execution match for Tool Use. A response counts as correct only when it matches the checker's target exactly --- a gold string or accepted paraphrase for General Skill Use and Cultural Skill Use, or the correct function name with matching argument slots for Tool Use. Checker acceptance criteria for each domain are detailed in Appendix~\ref{sec:app-checkers}; all other implementation details are in Appendix~\ref{sec:app-impl}.

Two tiers of models produce and score the main results in this paper. Tier 1 (skill generation) uses Qwen3.5-9B for both the self-generated skill layer and MT translation, representative of the lightweight, practical LLMs multilingual data-synthesis pipelines typically build on today \citep{liu2026midb,zhao2026mdaq} (\emph{e.g.}, MIDB adopts an 8B LLM as its main data synthesizer). Tier 2 uses Gemini-3-Flash as the scorer, router and solver, chosen for its balance of multilingual quality and speed, a trade-off widely adopted in recent multilingual agent research \citep{liu2026made,liu2026gaoyao}. See a backbone robustness analysis in Fig.~\ref{fig:backbone}.

\subsection{Experimental Results}

\paragraph{Main Results.}

Table~\ref{tab:main} shows M-SQE first or tied-first against the single strongest baseline in all 9 domain-by-retriever settings, reaching 69.1\% average success, +3.5pp over the strongest baseline, SkillFlow (65.6\%); significance analyses are in Appendix~\ref{sec:app-significance}. M-SQE is also strictly ahead of both selection anchors in all 9 settings (+6.3pp over Retrieve-only and +14.0pp over Random on average), confirming the premise of Section~\ref{sec:intro}: in realistic multilingual pools, what retrieval surfaces is often not what an agent can use, and scoring both views recovers the difference.
\begin{table}[b]
\centering\small
\setlength{\tabcolsep}{4pt}
\begin{tabular}{lccc}
\toprule
Method & General & Tool-Use & Cultural \\
\midrule
Theory-only & 67.0 & 28.7 & 59.6 \\
Action-only & 71.3 & 38.5 & 88.5 \\
\textbf{M-SQE} & \textbf{74.5} & \textbf{39.6} & \textbf{90.4} \\
\bottomrule
\end{tabular}
\caption{Mechanism ablation (BM25 retriever, Top3): task success rate (\%) for single view alone and full M-SQE.}
\label{tab:ablation}
\end{table}

\paragraph{Robustness across Regions and Languages.}

Fig.~\ref{fig:equity} shows M-SQE matches or beats both baselines in all six culture regions and 7 languages, with the largest gains on the benchmark's lowest-resource languages: Tool Use success rises +12.9pp on Hindi and +5.6pp on Swahili. These are the two languages whose in-language skill supply Table~\ref{tab:audit} measures at zero, so the improvement of M-SQE relieves exactly where the ecosystem leaves speakers with the least.

\paragraph{Ablation Study.}

Table~\ref{tab:ablation} isolates the Theory view and the Action view: M-SQE beats both single views on all three domains, though the two single-view gaps are visibly uneven. The unevenness follows from how the dimensions divide: among the quality dimensions we design, the skill-specific ones concentrate in the Action view (\emph{e.g.}, procedure match and constraint match), so a candidate's fit to the task weighs heavily on downstream success. The Theory view complements this as an intrinsic-quality guard, protecting the agent from the occasional catastrophically flawed skill that nonetheless reads well-fitting (as revealed by the empty-guidance Case 2 in Appendix~\ref{sec:app-cases}). Together, the two views cover complementary failure modes of multilingual agentic skills.

\begin{figure}[t]
\centering
\includegraphics[width=0.75\linewidth]{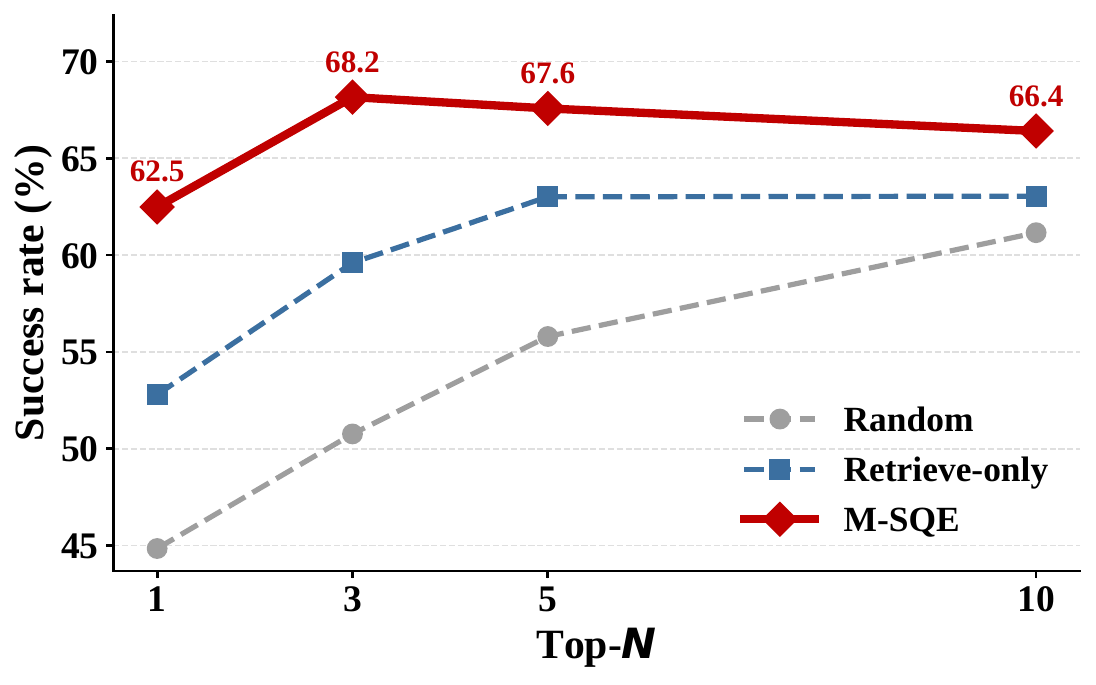}
\caption{Three-domain average task success vs.\ skill budget $N$ under BM25: M-SQE beats Retrieve-only at every budget, with the margin widening as the budget tightens.}
\label{fig:topn}
\end{figure}

\begin{figure}[t]
\centering
\includegraphics[width=0.8\linewidth]{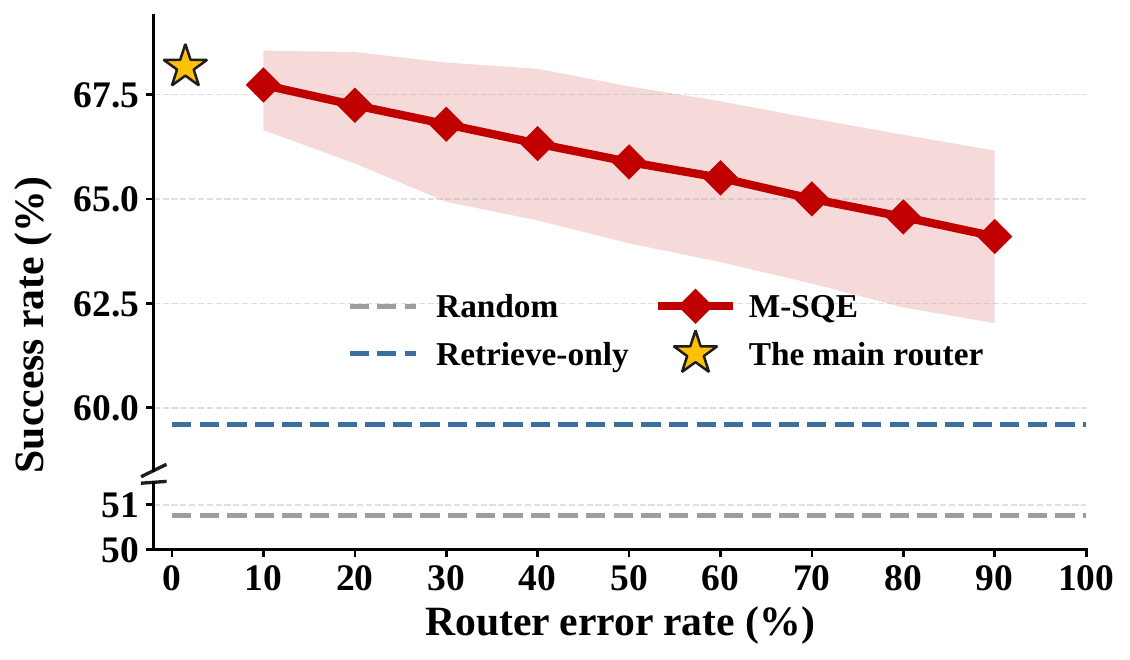}
\caption{Router corruption sweep (BM25, Top-3, three-domain average): M-SQE stays above both baselines at every corruption rate up to 90\% (1,000 seeds per rate; shaded band = empirical 95\% interval); the star marks the 5-shot router used in the main experiments with 1.5\% error rate.}
\label{fig:router}
\end{figure}

\paragraph{Skill Budget Sensitivity.} Fig.~\ref{fig:topn} shows M-SQE beats Retrieve-only at every skill budget $N \in \{1,3,5,10\}$, with the margin widening from +3.4pp at Top10 to +9.7pp at Top1. As the budget tightens, a bad skill does more damage to the solver (\emph{e.g.}, $N=1$ leaves no chance for a single bad pick), exactly where careful selection pays off. Small budgets are also the regime real agent deployments occupy: carrying ten skills can consume ten times the context of carrying one.

\paragraph{Routing Robustness.}

The domain router supplies the only input M-SQE requires beyond the query and its candidate skills; randomly flipping its predicted domains at rates from 10\% to 90\% (Fig.~\ref{fig:router}), M-SQE stays above Retrieve-only throughout, from +8.1pp at 10\% corruption to +4.5pp at 90\%. Even with the routing signal nearly destroyed, quality estimation alone keeps M-SQE ahead: routing sharpens the margin, and the two views hold the floor.

\begin{figure}[t]
\centering
\includegraphics[width=\linewidth]{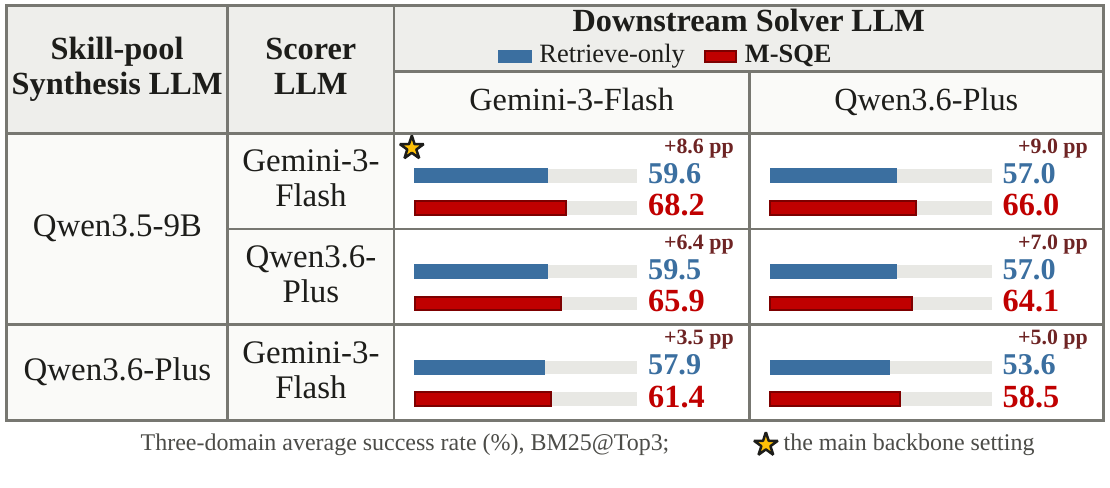}
\caption{M-SQE remains ahead of Retrieve-only across all six skill-pool synthesis, scorer, and downstream-solver backbone configurations.}
\label{fig:backbone}
\end{figure}

\paragraph{LLM Backbone Robustness.} To test whether M-SQE depends on a particular pipeline backbone, we vary the pool-synthesis, scorer, and downstream-solver LLMs. Across all six combinations, M-SQE remains +3.5--9.0pp above Retrieve-only (Fig.~\ref{fig:backbone}). Changing either the scorer or solver preserves this ordering, and rebuilding the MT and self-generated pool layers with a stronger Qwen3.6-Plus retains positive gains for both solvers. Together, these controls demonstrate that M-SQE's two-view quality estimation captures transferable skill utility across the pipeline, while ruling out pool-synthesis artifacts and scorer or solver bias as explanations for the gains.

\begin{table}[t]
\centering\small
\setlength{\tabcolsep}{6pt}
\begin{tabular}{lccc}
\toprule
Method & General & Tool-Use & Cultural \\
\midrule
Random & 36.17 & 12.08 & 42.31 \\
Retrieve-only & 36.17 & 12.83 & \textbf{50.00} \\
Base (untrained) & 35.11 & 10.19 & 38.46 \\
\textbf{M-SQE} & \textbf{38.30} & \textbf{13.21} & \textbf{50.00} \\
\bottomrule
\end{tabular}
\caption{Downstream trajectory generalization: task success rate (\%) of a Qwen3.5-9B model fine-tuned on trajectories built from each method's selected skills.}
\label{tab:downstream}
\end{table}

\paragraph{Downstream Trajectory Generalization.} Despite strong performance of M-SQE in aiding skill use, we further veirify its usefulness as training material for a downstream agent, and test whether skill quality still matters for a small open-source model. Fine-tuning Qwen3.5-9B as an agentic solver with trajectories built from each method's selected skills, M-SQE yields the best fine-tuned model averagely in three domains (Table~\ref{tab:downstream}). The signal that picks better skills to read also picks better skills to learn from, extending M-SQE from an inference-time filter to a beneficial training-data constructor.

\section{Conclusion}
\label{sec:conclusion}

In this paper, we introduced M-SQE, a post-retrieval quality estimation layer that judges a retrieved skill on its own merits and on its fit to the query, then routes the two judgments by the task's domain, turning the multilingual skill pool into one an agent can trust across domain and languages. As agentic AI reaches every language community, we hope M-SQE marks a step from agents that merely sound fluent toward agents that are equally capable in any language or culture. Future work includes extending to more languages and cultural regions, scaling to larger skill ecosystems, and broadening from skills to more agentic assets. Limitations and responsible use are discussed in Appendix~\ref{sec:app-limitations}.

\bibliography{msqe}

@article{xbcp2026,
  title={Beyond Monolingual Deep Research: Evaluating Agents and Retrievers with Cross-Lingual BrowseComp-Plus},
  author={Lu, Yuheng and Zeng, Qingcheng and Qi, Heli and Yu, Puxuan and Zhao, Fuheng and Yang, Rui and Yanaka, Hitomi and Yokoya, Naoto and Xuan, Weihao},
  journal={arXiv preprint arXiv:2606.15345},
  year={2026}
}

@article{skillsbench2026,
  title={SkillsBench: Benchmarking How Well Agent Skills Work Across Diverse Tasks},
  author={Li, Xiangyi and Liu, Yimin and Chen, Wenbo and You, Bingran and Di, Zonglin and He, Yifeng and Zheng, Shenghan and Choe, Kyoung Whan and Sun, Jiankai and Wang, Shuyi and others},
  journal={arXiv preprint arXiv:2602.12670},
  year={2026}
}

@article{skillflow2025,
  title={SkillFlow: Scalable and Efficient Agent Skill Retrieval System},
  author={Li, Fangzhou and Tagkopoulos, Pagkratios and Tagkopoulos, Ilias},
  journal={arXiv preprint arXiv:2504.06188},
  year={2025}
}

@inproceedings{liu2026midb,
  title={MIDB: Multilingual Instruction Data Booster for Enhancing Cultural Equality in Multilingual Instruction Synthesis},
  author={Liu, Yilun and Zhao, Chunguang and Yang, Xinhua and Zeng, Hongyong and Tao, Shimin and Meng, Weibin and He, Minggui and Yu, Yan and Ma, Hongxia and Zhang, Li and Wei, Daimeng and Chen, Boxing},
  booktitle={Proceedings of the AAAI Conference on Artificial Intelligence},
  volume={40},
  pages={38952--38961},
  doi={10.1609/aaai.v40i45.41241},
  url={https://ojs.aaai.org/index.php/AAAI/article/view/41241},
  year={2026}
}

@article{robertson2009bm25,
  title={The Probabilistic Relevance Framework: BM25 and Beyond},
  author={Robertson, Stephen and Zaragoza, Hugo},
  journal={Foundations and Trends in Information Retrieval},
  volume={3},
  number={4},
  pages={333--389},
  year={2009}
}

@inproceedings{karpukhin2020dpr,
  title={Dense Passage Retrieval for Open-Domain Question Answering},
  author={Karpukhin, Vladimir and Oguz, Barlas and Min, Sewon and Lewis, Patrick and Wu, Ledell and Edunov, Sergey and Chen, Danqi and Yih, Wen-tau},
  booktitle={Proceedings of the 2020 Conference on Empirical Methods in Natural Language Processing (EMNLP)},
  pages={6769--6781},
  year={2020}
}

@article{luan2021sparse,
  title={Sparse, Dense, and Attentional Representations for Text Retrieval},
  author={Luan, Yi and Eisenstein, Jacob and Toutanova, Kristina and Collins, Michael},
  journal={Transactions of the Association for Computational Linguistics},
  volume={9},
  pages={329--345},
  year={2021}
}

@inproceedings{rystrom2025multilingual,
  title={Multilingual != Multicultural: Evaluating Gaps Between Multilingual Capabilities and Cultural Alignment in {LLM}s},
  author={Rystr{\o}m, Jonathan Hvithamar and Kirk, Hannah Rose and Hale, Scott},
  booktitle={Proceedings of the Interdisciplinary Workshop on Observations of Misunderstood, Misguided and Malicious Use of Language Models},
  address={Varna, Bulgaria},
  publisher={INCOMA Ltd., Shoumen, Bulgaria},
  pages={74--85},
  year={2025}
}

@inproceedings{joshi2020state,
  title={The State and Fate of Linguistic Diversity and Inclusion in the NLP World},
  author={Joshi, Pratik and Santy, Sebastin and Budhiraja, Amar and Bali, Kalika and Choudhury, Monojit},
  booktitle={Proceedings of the 58th Annual Meeting of the Association for Computational Linguistics},
  pages={6282--6293},
  year={2020}
}

@article{stanford2025digitaldivide,
  author={Shana Lynch},
  title={How AI is leaving non-English speakers behind},
  journal={Stanford News},
  year={2025},
  month={May},
  url={https://news.stanford.edu/stories/2025/05/digital-divide-ai-llms-exclusion-non-english-speakers-research}
}

@article{wang2023voyager,
  title={Voyager: An Open-Ended Embodied Agent with Large Language Models},
  author={Wang, Guanzhi and Xie, Yuqi and Jiang, Yunfan and Mandlekar, Ajay and Xiao, Chaowei and Zhu, Yuke and Fan, Linxi and Anandkumar, Anima},
  journal={Transactions on Machine Learning Research},
  year={2024},
  url={https://openreview.net/forum?id=ehfRiF0R3a}
}

@article{xu2026agentskills,
  title={Agent Skills for Large Language Models: Architecture, Acquisition, Security, and the Path Forward},
  author={Xu, Renjun and Yan, Yang},
  journal={arXiv preprint arXiv:2602.12430},
  year={2026}
}

@article{zhou2026agentskillssurvey,
  title={A Comprehensive Survey on Agent Skills: Taxonomy, Techniques, and Applications},
  author={Zhou, Yingli and Wang, Shu and Su, Yaodong and Du, Wenchuan and Fang, Yixiang and Lin, Xuemin},
  journal={arXiv preprint arXiv:2605.07358},
  year={2026}
}

@article{zhang2026coevoskills,
  title={CoEvoSkills: Self-Evolving Agent Skills via Co-Evolutionary Verification},
  author={Zhang, Hanrong and Fan, Shicheng and Zou, Henry Peng and Chen, Yankai and Wang, Zhenting and Zhou, Jiayu and Li, Chengze and Huang, Wei-Chieh and Yao, Yifei and Zheng, Kening and Liu, Xue and Li, Xiaoxiao and Yu, Philip S.},
  journal={arXiv preprint arXiv:2604.01687},
  year={2026}
}

@inproceedings{zheng2024toolrerank,
  title={ToolRerank: Adaptive and Hierarchy-Aware Reranking for Tool Retrieval},
  author={Zheng, Yuanhang and Li, Peng and Liu, Wei and Liu, Yang and Luan, Jian and Wang, Bin},
  booktitle={Proceedings of the 2024 Joint International Conference on Computational Linguistics, Language Resources and Evaluation (LREC-COLING 2024)},
  publisher={ELRA and ICCL},
  pages={16263--16273},
  url={https://aclanthology.org/2024.lrec-main.1413/},
  year={2024}
}

@inproceedings{liu2024deita,
  title={What Makes Good Data for Alignment? A Comprehensive Study of Automatic Data Selection in Instruction Tuning},
  author={Liu, Wei and Zeng, Weihao and He, Keqing and Jiang, Yong and He, Junxian},
  booktitle={Proceedings of the Twelfth International Conference on Learning Representations (ICLR)},
  year={2024}
}

@article{bonifacio2021mmarco,
  title={mMARCO: A Multilingual Version of the MS MARCO Passage Ranking Dataset},
  author={Bonifacio, Luiz and Jeronymo, Vitor and Abonizio, Hugo Queiroz and Campiotti, Israel and Fadaee, Marzieh and Lotufo, Roberto and Nogueira, Rodrigo},
  journal={arXiv preprint arXiv:2108.13897},
  year={2021}
}

@inproceedings{ali2025jql,
  title={Judging Quality Across Languages: A Multilingual Approach to Pretraining Data Filtering with Language Models},
  author={Ali, Mehdi and Brack, Manuel and L{\"u}bbering, Max and Wendt, Elias and Khan, Abbas Goher and Rutmann, Richard and Jude, Alex and Kraus, Maurice and Weber, Alexander Arno and Stollenwerk, Felix and Kacz{\'e}r, David and Mai, Florian and Flek, Lucie and Sifa, Rafet and Flores-Herr, Nicolas and Koehler, Joachim and Schramowski, Patrick and Fromm, Michael and Kersting, Kristian},
  booktitle={Proceedings of the 2025 Conference on Empirical Methods in Natural Language Processing},
  pages={8859--8898},
  year={2025}
}

@inproceedings{lai2024llms,
  title={{LLM}s Beyond {E}nglish: Scaling the Multilingual Capability of {LLM}s with Cross-Lingual Feedback},
  author={Lai, Wen and Mesgar, Mohsen and Fraser, Alexander},
  booktitle={Findings of the Association for Computational Linguistics: ACL 2024},
  address={Bangkok, Thailand},
  publisher={Association for Computational Linguistics},
  doi={10.18653/v1/2024.findings-acl.488},
  pages={8186--8213},
  year={2024}
}

@inproceedings{zhao2026mdaq,
  title={M-DaQ: Retrieving Samples with Multilingual Diversity and Quality for Instruction Fine-Tuning Datasets},
  author={Zhao, Chunguang and Liu, Yilun and Zeng, Pufan and Luo, Yuanchang and Tao, Shimin and He, Minggui and Meng, Weibin and Xu, Song and Liu, Chen and Ma, Hongxia and Zhang, Li and Chen, Boxing and Wei, Daimeng},
  booktitle={Proceedings of the 49th International ACM SIGIR Conference on Research and Development in Information Retrieval},
  pages={4361--4366},
  year={2026},
  doi={10.1145/3805712.3809946}
}

@inproceedings{kulkarni2025massiveagents,
  title={MASSIVE-Agents: A Benchmark for Multilingual Function-Calling in 52 Languages},
  author={Kulkarni, Mayank and Mazzia, Vittorio and Gaspers, Judith and Hench, Chris and FitzGerald, Jack},
  booktitle={Findings of the Association for Computational Linguistics: EMNLP 2025},
  publisher={Association for Computational Linguistics},
  pages={20193--20215},
  doi={10.18653/v1/2025.findings-emnlp.1099},
  url={https://aclanthology.org/2025.findings-emnlp.1099/},
  year={2025}
}

@inproceedings{rao2024normad,
  title={NormAd: A Framework for Measuring the Cultural Adaptability of Large Language Models},
  author={Rao, Abhinav and Yerukola, Akhila and Shah, Vishwa and Reinecke, Katharina and Sap, Maarten},
  booktitle={Proceedings of the 2025 Conference of the Nations of the Americas Chapter of the Association for Computational Linguistics: Human Language Technologies (Volume 1: Long Papers)},
  address={Albuquerque, New Mexico},
  publisher={Association for Computational Linguistics},
  pages={2373--2403},
  doi={10.18653/v1/2025.naacl-long.120},
  url={https://aclanthology.org/2025.naacl-long.120/},
  year={2025}
}

@inproceedings{shi2024culturebank,
  title={CultureBank: An Online Community-Driven Knowledge Base Towards Culturally Aware Language Technologies},
  author={Shi, Weiyan and Li, Ryan and Zhang, Yutong and Ziems, Caleb and Yu, Sunny and Horesh, Raya and Paula, Rog{\'e}rio Abreu De and Yang, Diyi},
  booktitle={Findings of the Association for Computational Linguistics: EMNLP 2024},
  pages={4996--5025},
  doi={10.18653/v1/2024.findings-emnlp.288},
  url={https://aclanthology.org/2024.findings-emnlp.288/},
  year={2024}
}

@article{lin2026culturall,
  title={CulturALL: Benchmarking Multilingual and Multicultural Competence of LLMs on Grounded Tasks},
  author={Lin, Peiqin and Lyu, Chenyang and Luo, Wenjiang and Ye, Haotian and Hossain, Md Mehrab and Ma, Chunlan and Ji, Shaoxiong and Samih, Younes and Zeng, Bo and Jiang, Fan and Cao, Yuanbin and Duisenbek, Dilda and Xun, Adrian Neo Sau and Pozdniakova, Daria and Misevich, Liubou and Marinkovi{\'c}, Nevena and Nguyen, Ngoc Gia Linh and Do, Thi Khanh Linh and Sophy, Sarakmatak and Hu, Baotian and Chen, Guanhua and Tang, Gongbo and Aji, Alham Fikri and Wang, Longyue and Luo, Weihua},
  journal={arXiv preprint arXiv:2604.19262},
  year={2026}
}

@article{zhang2025culturescope,
  title={CultureScope: A Dimensional Lens for Probing Cultural Understanding in LLMs},
  author={Zhang, Jinghao and Jiang, Sihang and Guo, Shiwei and Chen, Shisong and Xiao, Yanghua and Feng, Hongwei and Liang, Jiaqing and He, Minggui and Tao, Shimin and Ma, Hongxia},
  journal={arXiv preprint arXiv:2509.16188},
  year={2025}
}

@article{fung2024cultureatlas,
  title={No Culture Left Behind: Massively Multi-Cultural Knowledge Acquisition \& LM Benchmarking on 1000+ Sub-Country Regions and 2000+ Ethnolinguistic Groups},
  author={Fung, Yi R. and Zhao, Ruining and Doo, Jae and Sun, Chenkai and Ji, Heng},
  journal={arXiv preprint arXiv:2402.09369},
  year={2024}
}

@article{guo2026sage,
  title={Do Large Language Models Truly Understand Cross-Cultural Differences?},
  author={Guo, Shiwei and Jiang, Sihang and He, Qianxi and Xiao, Yanghua and Liang, Jiaqing and Bi, Yude and He, Minggui and Tao, Shimin and Zhang, Li},
  journal={arXiv preprint arXiv:2512.07075},
  year={2025}
}

@misc{littledinoc2026agentskills,
  title={agent-skills},
  author={{LittleDinoC}},
  howpublished={Hugging Face Datasets},
  year={2026},
  url={https://huggingface.co/datasets/LittleDinoC/agent-skills},
  note={Accessed 2026-07-24}
}

@misc{wshobson2026agents,
  title={agents: A Multi-Harness Agentic Plugin Marketplace},
  author={{wshobson}},
  howpublished={GitHub repository},
  year={2026},
  url={https://github.com/wshobson/agents},
  note={Accessed 2026-07-24}
}

@misc{huzey2026claudeskills,
  title={claude-skills},
  author={{huzey}},
  howpublished={Hugging Face Datasets},
  year={2026},
  url={https://huggingface.co/datasets/huzey/claude-skills},
  note={Accessed 2026-07-24}
}

@article{liu2026made,
  title={MADE: Beyond Scoring via a Multilingual Agentic Diagnosing Engine for Fine-Grained Evaluation Insights},
  author={Liu, Yilun and Zhang, Miao and Tao, Shimin and He, Minggui and Zhao, Chunguang and Liu, Chenxin and Zhang, Li and Liu, Chen and Qian, Cheng and Deng, Liqun and Meng, Xiaojun and Wei, Daimeng},
  journal={arXiv preprint arXiv:2606.07020},
  year={2026}
}

@inproceedings{liu2026gaoyao,
  title={The {G}ao{Y}ao Benchmark: A Comprehensive Framework for Evaluating Multilingual and Multicultural Abilities of Large Language Models},
  author={Liu, Yilun and Zhao, Chunguang and Piao, Mengyao and Miao, Lingqi and Tao, Shimin and He, Minggui and Liu, Chenxin and Zhang, Li and Ma, Hongxia and Guo, Jiaxin and Liu, Chen and Deng, Liqun and Wei, Jiansheng and Meng, Xiaojun and Du, Fanyi and Wei, Daimeng and Xiao, Yanghua},
  booktitle={Proceedings of the 64th Annual Meeting of the Association for Computational Linguistics (Volume 1: Long Papers)},
  address={San Diego, California, United States},
  publisher={Association for Computational Linguistics},
  doi={10.18653/v1/2026.acl-long.977},
  pages={21364--21384},
  year={2026}
}

@inproceedings{long2024llmsyndata,
  title={On {LLM}s-Driven Synthetic Data Generation, Curation, and Evaluation: A Survey},
  author={Long, Lin and Wang, Rui and Xiao, Ruixuan and Zhao, Junbo and Ding, Xiao and Chen, Gang and Wang, Haobo},
  booktitle={Findings of the Association for Computational Linguistics: ACL 2024},
  address={Bangkok, Thailand},
  publisher={Association for Computational Linguistics},
  doi={10.18653/v1/2024.findings-acl.658},
  pages={11065--11082},
  year={2024}
}

@article{ma2026skillgen,
  title={SkillGen: Verified Inference-Time Agent Skill Synthesis},
  author={Ma, Yuchen and Huang, Yue and Bao, Han and Zhuang, Haomin and Shukla, Swadheen and Galley, Michel and Zhang, Xiangliang and Feuerriegel, Stefan},
  journal={arXiv preprint arXiv:2605.10999},
  year={2026}
}

@article{wang2026skillx,
  title={SkillX: Automatically Constructing Skill Knowledge Bases for Agents},
  author={Wang, Chenxi and Yu, Zhuoyun and Xie, Xin and Yao, Wuguannan and Fang, Runnan and Qiao, Shuofei and Cao, Kexin and Zheng, Guozhou and Qi, Xiang and Zhang, Peng and Deng, Shumin},
  journal={arXiv preprint arXiv:2604.04804},
  year={2026}
}

@inproceedings{hu2022lora,
  title={{LoRA}: Low-Rank Adaptation of Large Language Models},
  author={Hu, Edward J. and Shen, Yelong and Wallis, Phillip and Allen-Zhu, Zeyuan and Li, Yuanzhi and Wang, Shean and Wang, Lu and Chen, Weizhu},
  booktitle={International Conference on Learning Representations},
  year={2022},
  url={https://openreview.net/forum?id=nZeVKeeFYf9}
}

\clearpage
\appendix
\section{Community Skill Ecosystem Audit}
\label{sec:app-audit}

Section~\ref{sec:intro} reports that non-English agent skills are scarce and, for several widely spoken languages, effectively absent from the public ecosystem (Table~\ref{tab:audit}). This section describes the audit protocol behind that claim and reports the full per-language methodological detail.

\paragraph{Audit scope.} We audited over a dozen public community skill indexes and repositories on code- and dataset-sharing platforms such as Hugging Face and GitHub, contributing $\sim$84,700 entries in total (un-deduplicated across sources), the figure Table~\ref{tab:audit}'s counts are computed against. The largest sources are LittleDinoC's \texttt{agent-skills} collection on Hugging Face Datasets ($\sim$61,000 entries) \citep{littledinoc2026agentskills}, wshobson's \texttt{agents} multi-harness agentic plugin marketplace on GitHub (175 skills, a 38.2k-star repository) \citep{wshobson2026agents}, and huzey's \texttt{claude-skills} collection on Hugging Face Datasets (22,862 curated SKILL.md files crawled from skills.sh across 522 source repositories) \citep{huzey2026claudeskills}; the remaining repositories were surfaced by these indexes and our own targeted per-language searches. All were active, community-maintained collections at the time of the audit (2026-07) -- the two Hugging Face collections each aggregate skills crawled from a large number of independent GitHub repositories, and the GitHub marketplace is itself a widely-starred skill repository in its own right -- so together they approximate the pool an agent's skill retriever would plausibly draw from in practice. The goal was to estimate, for each of the paper's evaluated languages, how many entries in this ecosystem carry a skill body that is genuinely authored in the target language, as opposed to an English skill with translated metadata, an MT stub, or a title that merely contains a target-language keyword. The English count in Table~\ref{tab:audit} is the audited total minus all identified in-language content; entries whose bodies are MT stubs wrapped around English instructions count toward the English side.

\paragraph{Method.} We used Unicode-script-based in-language detection: the presence of kana characters flags Japanese, hangul flags Korean, and CJK ideographs without kana or hangul flag Chinese. Script matches below a 25-character threshold were discarded, since short matches are dominated by false positives — English-language skill bodies whose title, tags, or embedded example strings happen to contain a target-language trigger word. Every repository that survived this script filter was then fetched directly and its skill body checked, programmatically and by hand, to confirm genuine in-language prose content rather than a templated header, a boilerplate translation notice, or a partial MT wrapped around otherwise-English instructions.

Table~\ref{tab:audit} reports the resulting counts. The script screen and length threshold apply to every audited source; the precision of each count follows its flagged volume. Chinese surfaces by far the most flagged entries, so its genuine in-language count is estimated from expert spot-checks of the flagged set; because skill authorship clusters by author (a few prolific contributors account for a disproportionate share of any one language), we read this estimate at order-of-magnitude precision. French, Korean, and Japanese each surface only a few hundred flagged candidates, so every flagged repository was checked directly and their counts are rounded tallies. The zero counts for Swahili and Hindi come from direct enumeration of every candidate repository the script filter and targeted keyword search surfaced, each verified individually.

Even the best-represented non-English language, Chinese, accounts for only a small fraction of the $\sim$84,700-entry combined index, and the count falls off sharply for French, Korean, and Japanese. For Swahili and Hindi — languages spoken natively by hundreds of millions of people combined — we found no in-language skill content at all. This scarcity is the reason a realistic multilingual skill pool cannot be built from ecological material alone. Section~\ref{sec:app-pool} describes how each evaluation domain's skill pool accounts for this by combining ecological-style skills with MT and self-generated material, mirroring the de facto composition of the ecosystem this audit documents.

\section{Statistical Significance}
\label{sec:app-significance}

Table~\ref{tab:app-significance} compares M-SQE with Retrieve-only by skill-use domain, retriever, and downstream skill budget. To avoid treating repeated evaluations of the same task as independent, we first average outcomes within each task and then compute paired bootstrap intervals and sign-flip $p$-values over tasks.

\begin{table}[t]
\centering
\setlength{\tabcolsep}{3pt}
\resizebox{\columnwidth}{!}{%
\begin{tabular}{@{}lccc@{}}
\toprule
Slice & $\Delta$ vs Retrieve (pp) & 95\% CI & task-level $p$ \\
\midrule
\multicolumn{4}{l}{\emph{Domain-level}} \\
General Skill Use & +4.3 & [+1.0, +7.6] & 0.0158* \\
Tool Use & +6.0 & [+4.1, +8.1] & $<$0.0001*** \\
Cultural Skill Use & +5.0 & [+1.1, +9.3] & 0.0213* \\
\midrule
\multicolumn{4}{l}{\emph{Retriever-level}} \\
BM25 & +6.6 & [+4.2, +9.0] & $<$0.0001*** \\
Neural & +7.7 & [+5.3, +10.3] & $<$0.0001*** \\
SkillFlow & +2.1 & [+0.6, +3.7] & 0.0079** \\
\midrule
\multicolumn{4}{l}{\emph{Budget-level}} \\
Top1 & +7.7 & [+5.2, +10.3] & $<$0.0001*** \\
Top3 & +7.1 & [+4.7, +9.6] & $<$0.0001*** \\
Top5 & +4.1 & [+2.1, +6.2] & 0.0002*** \\
Top10 & +2.9 & [+1.3, +4.6] & 0.0009*** \\
\bottomrule
\end{tabular}
}
\caption{Paired M-SQE vs.\ Retrieve-only tests. Domain and retriever rows summarize all four skill budgets; budget rows hold TopN fixed. To avoid treating repeated evaluations of the same task as independent, each row first averages repeated outcomes within task, then reports a paired sign-flip $p$-value and task-level bootstrap 95\% CI from 10{,}000 resamples; \emph{*} $p{<}0.05$, \emph{**} $p{<}0.01$, \emph{***} $p{<}0.001$.}
\label{tab:app-significance}
\end{table}

\paragraph{Domain-level.} M-SQE improves over Retrieve-only in General Skill Use (+4.3pp), Tool Use (+6.0pp), and Cultural Skill Use (+5.0pp). All three task-level tests are significant ($p=0.0158$, $p<0.0001$, and $p=0.0213$, respectively), with bootstrap intervals that exclude zero.

\paragraph{Retriever-level.} The gain holds under BM25 (+6.6pp), Neural (+7.7pp), and SkillFlow (+2.1pp). The task-level tests remain significant for all three retrievers ($p<0.0001$, $p<0.0001$, and $p=0.0079$), showing that the effect is not tied to one candidate-generation mechanism.

\paragraph{Budget-level.} M-SQE improves over Retrieve-only by +7.7pp at Top1, +7.1pp at Top3, +4.1pp at Top5, and +2.9pp at Top10. All four task-level tests are significant ($p<0.0001$, $p<0.0001$, $p=0.0002$, and $p=0.0009$), confirming that the gain persists across the full budget range while narrowing as more retrieved skills are passed to the solver.

\section{Hyperparameter Stability Analysis}
\label{sec:app-stability}

This section asks whether M-SQE's combination constants (Section~\ref{sec:method}) are knife-edged, that is, whether a small nudge to any one of them would already change which candidates the paper's selector hands to the downstream solver. Two of the three fusion rules carry exactly one tunable scalar, and both were fixed by the design rationale given in Section~\ref{sec:method} rather than by fitting any evaluation outcome: the General fusion's mixing weight $\lambda=0.6$ gives Action a modest majority over Theory ($C_{\text{gen}} = 0.6\,\text{Action} + 0.4\,\text{Theory}$), and the Tool-Use guard's threshold $\tau=65$ marks the point on the 0--100 Theory scale below which a candidate is judged structurally unusable regardless of how relevant it looks. The Cultural fusion has no comparable free scalar to begin with: $C_{\text{cul}} = z(\text{Retrieval}) + z(\text{Theory}) + z(\text{Action})$ sums the three per-query standardized views with equal (1:1:1) weight by construction, not a tuned coefficient. We nonetheless subject this equal weighting to the same stability treatment below, asking whether nudging any one view's weight away from parity would already change the selection. What follows verifies, empirically and per task, that none of these three commitments sits on a knife edge.

Concretely, for every task we define its \emph{stability width} as the length of the interval of parameter values over which that task's selected Top-3 set stays completely unchanged compared with Table~\ref{tab:main}. A stability width under 0.02 for $\lambda$, under 2 points for $\tau$, or under 0.04 for a Cultural fusion weight (on the reference scale defined below) means the deployed value sits close enough to a boundary that a small nudge --- respectively $\pm0.01$, $\pm1$ point, or $\pm0.02$ --- already flips which three skills are selected for that task; this is the operational meaning of ``knife-edge'' throughout this section.

\paragraph{General Skill Use.} Across the 94 tasks (BM25 retriever), the per-task stability width for $\lambda$ has a median of 0.635 and a mean of 0.656 over the full $[0,1]$ range $\lambda$ can take; one task (1/94, 1.1\%) is stable for every $\lambda\in[0,1]$, its Theory/Action candidate ranking never crossing regardless of the weight. No task (0/94, 0.0\%) has a stability width under 0.02, and no task sits at an exact tie with a non-selected candidate at $\lambda=0.6$ itself (0/94). The selected Top-3 is identical across a $\pm0.05$ sweep of the mixing weight (0.55--0.65) in 83/94 (88.3\%) of tasks, and across a $\pm0.10$ sweep (0.50--0.70) in 79/94 (84.0\%).

\paragraph{Tool Use.} Across the 265 tasks (BM25 retriever), the per-task stability width for $\tau$ has a median of 95 and a mean of 96 points over the full $[0,100]$ Theory scale; 80/265 (30.2\%) of tasks are stable for every $\tau\in[0,100]$. No task (0/265, 0.0\%) has a stability width under 2 points. The Theory$\geq$65 guard's selection is identical across a $\pm5$-point sweep of the guard threshold (60--70) in 265/265 (100.0\%) of tasks, and across a $\pm10$-point sweep (55--75) in 265/265 (100.0\%).

\paragraph{Cultural Skill Use.} Since $C_{\text{cul}}$ carries no single scalar to sweep, we instead perturb each of its three equal weights one at a time, holding the other two fixed at 1, across the 52 tasks (BM25 retriever); a reference domain of $[0,2]$ (0 dropping a view entirely, 2 double-counting it) stands in for the $[0,1]$ and $[0,100]$ ranges $\lambda$ and $\tau$ naturally live in. No task has a stability width under 0.04 for any of the three views --- zero knife-edge tasks for Retrieval, Theory, or Action individually. A $\pm0.05$ nudge to a single view's weight leaves the Top-3 unchanged in 94.2\% (49/52) of tasks for Retrieval, 92.3\% (48/52) for Theory, and 96.2\% (50/52) for Action; at $\pm0.10$ these fall to 82.7\% (43/52), 86.5\% (45/52), and 92.3\% (48/52); at $\pm0.25$ to 57.7\% (30/52), 69.2\% (36/52), and 71.2\% (37/52). Requiring simultaneous stability under all three single-weight nudges --- the strictest reading, in which a task counts only if none of the three views flips its selection --- still leaves 92.3\% (48/52) of tasks unchanged at $\pm0.05$, 80.8\% (42/52) at $\pm0.10$, and 48.1\% (25/52) at $\pm0.25$. The equal standardized weighting therefore sits inside as broad a stability plateau as the two tuned constants above, despite carrying no free scalar to tune in the first place.

Across all three domains, the deployed constants --- two fixed by design rationale, one fixed by construction --- lie within broad selection-stability plateaus for the large majority of tasks rather than on a knife edge.

\begin{table*}[t]
\centering\small
\begin{tabular}{p{2.4cm} p{5.6cm} p{7.6cm}}
\toprule
Dimension & Explanation & Skill-side judgment case \\
\midrule
Correctness [red-line] & The skill's procedural facts and logic are right; missing external files or tools is not itself a correctness issue. & An \texttt{XLOOKUP} skill instructs the reader to leave the fourth argument blank for an exact-match lookup, when the target spreadsheet application requires it set to 0 -- a wrong instruction that silently produces a broken formula. \\
Completeness [basic] & The necessary steps and boundary cases for the task class are covered, with no critical omission. & A PDF-form-filling skill documents populating interactive fields but never covers a form with no interactive fields at all, leaving the agent with no fallback for a boundary case it will routinely meet. \\
Executability [basic] & Steps are concrete enough for an agent to act on, not merely descriptive. & A smart-home skill tells the agent to ``adjust the appropriate device'' without naming the device-location slot or its expected value format, so no executable function call can be built from it as written. \\
Cross-lingual faithfulness [basic] & No meaning-altering mistranslation or untranslated fragments that would impede use in the target locale. & A MT Hindi skill inverts the logic of a conditional ``if'' clause during translation, an error that propagates into every response the skill produces. \\
Localization [basic] & The prose is genuinely written for the expected target locale, not merely translated into it. & A skill tagged as a Hindi smart-home guide has a translated title but a body written almost entirely in fluent English -- usable to a bilingual reader, but not a Hindi-facing skill. \\
Context efficiency [advanced] & Concise enough not to dilute the agent's attention; an excessively long or padded skill can itself cause execution failure. & A red-envelope custom skill opens with three paragraphs on the history of Lunar New Year before ever stating the customary gift amount the query actually needs. \\
\bottomrule
\end{tabular}
\caption{The six Theory-view dimensions (Section~\ref{sec:method}): detailed explanation and one concrete skill-side judgment case per dimension.}
\label{tab:app-theory-dims}
\end{table*}

\begin{table*}[t]
\centering\small
\begin{tabular}{p{2.4cm} p{5.6cm} p{7.6cm}}
\toprule
Dimension & Explanation & Skill-side judgment case \\
\midrule
Task applicability & Whether the skill targets this task at all. & For a query asking to build an \texttt{XLOOKUP} formula, a general ``Excel functions overview'' skill is topically related but never demonstrates \texttt{XLOOKUP} syntax, so it scores low despite the topical overlap. \\
Procedure match & Whether the skill's steps cover what the query actually requires. & A calendar skill's steps assume a single one-time event, but the query asks to schedule a weekly recurring meeting; the skill never covers the recurrence step the query needs. \\
Constraint match & Whether slots, parameters, and constraints line up with the request. & For the query ``turn off the bathroom light,'' a smart-home skill documents a device-control function but never states a room/location slot, so its documented constraints do not line up with what the request needs specified. \\
Output-format match & Whether the skill produces the output shape the task expects. & A tool-use query needs a single well-formed JSON function call; a candidate skill's worked examples all end in a natural-language confirmation sentence instead of the structured call the checker expects. \\
Language fit & Whether the prose is usable in the query's language. & A Swahili query is paired with a candidate skill whose instructions are written entirely in French; procedurally sound, but not usable to a reader who only reads the query's language. \\
Misleading risk & The chance a related-looking skill steers the solver to a wrong API, value, or convention. & A wedding-gift-etiquette skill retrieved for a Lunar New Year red-envelope question reads as fluent and on-topic but grounds its answer in the wrong occasion, risking a confident, wrong response. \\
\bottomrule
\end{tabular}
\caption{The six Action-view dimensions (Section~\ref{sec:method}): detailed explanation and one concrete skill-side judgment case per dimension.}
\label{tab:app-action-dims}
\end{table*}

\section{Dimension Construction and Scoring Prompt Templates}
\label{sec:app-prompts}

\paragraph{Expert team.} Both views' dimensions were built by a team of professional language-service experts averaging over six years of experience in translation, localization, and multilingual content editing, drawn from an international language-service organization and covering the paper's evaluation languages (French, Hindi, Japanese, Korean, Swahili, and Chinese) alongside English. Experts were allocated to languages by native proficiency, and every dimension decision was cross-checked by a second expert before being finalized, mirroring the task-allocation and review discipline MIDB \citep{liu2026midb} used to build its own instruction-data quality taxonomy.

\paragraph{Construction process.} The dimensions were derived empirically, following the same audit-driven process MIDB \citep{liu2026midb} used to build its own
 quality criteria. Starting from the $\sim$84,700-entry community skill audit (Section~\ref{sec:app-audit}), the expert team sampled per-language sub-pools of skills for manual review. These development subsets were strictly disjoint from both the skill pools and evaluation tasks used in all reported experiments, preventing leakage. A first round of blind scoring rated each sampled skill against a provisional checklist adapted from MIDB's instruction-data criteria. The team then compared the lowest- and highest-scored items and traced the gap to failure modes the provisional checklist missed: skills whose steps looked complete on paper but were not concrete enough for an agent to execute, skills padded with irrelevant detail that diluted an agent's attention, and relevant-looking skills that nonetheless steered a solver toward a wrong API, value, or convention. These gaps were distilled into the Theory view's executability and context-efficiency dimensions and the Action view's dimensions such as misleading-risk. Two further audit rounds refined dimension wording and cut-line examples until inter-expert agreement on a shared calibration subset stabilized, at which point both rubrics were frozen for use throughout this paper. Table~\ref{tab:app-theory-dims} and Table~\ref{tab:app-action-dims} give a detailed explanation and one concrete skill-side judgment case for each Theory and Action dimension.

Both the Theory view and the Action view score every retrieved candidate through a single LLM call that returns a structured JSON object. The Theory rubric's dimensions instantiate the expert-validated multilingual quality taxonomy of \citet{liu2026midb} (content, translation, and localization criteria) extended with the two agent-specific dimensions discussed in Section~\ref{sec:method}. Fig.~\ref{fig:app-theory-prompt} and Fig.~\ref{fig:app-action-prompt} display the canonical rubric for each view, taken from the General Skill Use domain. For each domain, \emph{e.g.}, the Tool Use and Cultural Skill Use, the rubric reuse the same evaluator framing and 0--100 JSON-object output contract, while substituting the serialized candidate content for the domain at hand — a function schema and required argument slots for Tool Use, a culture entity and target region for Cultural Skill Use. Placeholders (\texttt{\{task\}}, \texttt{\{skill\_body\}}, \texttt{\{skill\_title\}}, \texttt{\{locale\}}) stand in for the per-instance fields substituted at call time. The 5-shot-per-domain router prompt template is shown in Fig.~\ref{fig:app-router-prompt}.

\begin{figure*}[t]
\centering
{\small
\begin{verbatim}
SYSTEM: You are an expert evaluator of AGENT SKILLS (procedural documents an agent
retrieves, loads into context, and follows to perform a class of tasks). Judge the
skill's intrinsic QUALITY. Do NOT judge relevance to any query, and do NOT reward
verbosity. Score EACH dimension INDEPENDENTLY -- a flaw in one dimension must not
lower another. You are blind to who authored or translated the skill.

Expected target locale: {locale}. Judge whether the PROSE of the skill is written
for this expected target locale. Ignore code blocks, package/API names, file paths,
commands, and variable names when judging prose language -- keeping those in
English is normal and must NOT be penalized. Substantial English prose while the
expected target locale is non-English is a localization failure even if the
English itself is fluent.

Dimensions (each scored 0-100; "violated":true means it fails this dimension's bar):
- correctness [red-line]: procedural facts/logic are correct; missing external
  files/tools is NOT a correctness issue.
- executability [basic]: steps are concrete and executable by an agent; needed
  tools/files are defined or obtainable.
- completeness [basic]: covers the necessary steps and boundary cases for this
  task class; no critical omission.
- cross-lingual faithfulness [basic]: no meaning-altering mistranslation, no
  untranslated fragments that impede use in the expected target locale.
- localization [basic]: usable as a skill for the expected target locale; penalize
  wrong-language or mixed-language prose, awkward literal translation, and wrong
  locale conventions (code/API names/paths/commands are exempt).
- context-efficiency [advanced]: concise, not bloated; an excessively long or
  padded skill dilutes the agent's attention and can itself cause execution
  failure.

Every "score" MUST be an integer 0-100; "confidence" likewise 0-100. Return ONLY
a JSON object:
{"dimensions": {"<dim>": {"score": <0-100 int>, "violated": <true|false>,
  "reason": "<=12 words"}}, "root_cause_tags": ["..."], "confidence": <0-100 int>}

SKILL: {skill_body}
\end{verbatim}
}
\caption{Theory scoring prompt template (General Skill Use domain). The overall Theory score is a deterministic function of the returned dimension scores, computed outside the model call: the mean of the six dimension scores, capped at 40 if any red-line dimension is violated, at 80 if any basic dimension is violated (and no red-line violation occurred), or left uncapped at 100 otherwise — a single severe defect can therefore not be averaged away by otherwise-strong dimensions.}
\label{fig:app-theory-prompt}
\end{figure*}

\begin{figure*}[t]
\centering
{\small
\begin{verbatim}
SYSTEM: You are an action-oriented evaluator for retrieved agent skills. Return
ONLY a JSON object. Do not use markdown.

USER: Evaluate whether the CANDIDATE SKILL is expected to help an agent solve
THIS TASK. You see only the task and the candidate skill document; you are not
given the source skill id, provenance, gold answer, checker, previous answer, or
execution result.

Rules:
- Judge task-specific expected utility, not intrinsic writing quality.
- Relevance is not enough. Penalize related skills that can lead to a wrong API,
  field name, formula, step order, locale convention, output shape, or language
  output.
- High expected_utility requires strong applicability plus low misleading_risk.
- Do not use or mention whether the skill "looks like the original skill"; you
  cannot know that.
- Code/API/file names may stay in English in non-English tasks. Penalize
  language only when prose mismatch blocks task use.

Return exactly this JSON schema:
{"task_applicability": 0-100, "procedure_match": 0-100, "constraint_match": 0-100,
 "output_format_match": 0-100, "language_fit": 0-100, "misleading_risk": 0-100,
 "expected_utility": 0-100, "brief_reason": "<=25 words"}

TASK: {task}
CANDIDATE SKILL TITLE: {skill_title}
CANDIDATE SKILL BODY: {skill_body}
\end{verbatim}
}
\caption{Action scoring prompt template (General Skill Use domain). One scoring call returns the overall Action score (\texttt{expected\_utility}) together with the six dimension scores: the dimension scores walk the scorer through each fit check, and the overall score weighs them for the query at hand, permitted to be high only when the checks pass and the misleading risk is low. The aggregation differs from the Theory view's fixed mean-with-caps rule (Fig.~\ref{fig:app-theory-prompt}) because the two views judge different objects: Theory judges the skill in isolation, where its dimensions are fixed document properties and one rule fits every query, while Action judges the skill against the current query, where the dimension that matters most changes from query to query and no single fixed weighting exists.}
\label{fig:app-action-prompt}
\end{figure*}

\begin{figure*}[t]
\centering
{\small
\begin{verbatim}
SYSTEM: You are a lightweight task-type classifier for an agent system.
Classify the user's query into exactly one task type:

- general: stand-alone knowledge, reasoning, coding, document/data manipulation,
  or procedural tasks answered directly by the agent.
- function: select or invoke an API, tool, app action, or service function and
  fill its arguments.
- culture: answer centrally depends on culture-specific customs, values,
  etiquette, social norms, history, or regional practices.

Tie rules: choose culture when culture-specific knowledge is essential. Choose
function only for an app/tool/service action or intent invocation; an ordinary
programming question that mentions a function is general. Otherwise choose general.

Return exactly one JSON object: {"tau":"general|function|culture"}.

FEW-SHOT EXAMPLES:
USER: [5 General Skill Use examples]
ASSISTANT: {"tau":"general"}
USER: [5 Tool Use examples]
ASSISTANT: {"tau":"function"}
USER: [5 Cultural Skill Use examples]
ASSISTANT: {"tau":"culture"}

USER: {query}
\end{verbatim}
}
\caption{Domain-router prompt template. Each bracketed placeholder denotes the five fixed demonstrations supplied for that domain.}
\label{fig:app-router-prompt}
\end{figure*}

\section{Pool Construction and Leakage Protocol}
\label{sec:app-pool}

Each evaluation domain draws its candidates from a three-layer skill pool. The three layers mirror the three ways a multilingual skill comes to exist given the ecosystem documented in Section~\ref{sec:app-audit}, and each preserves the artifacts its own production path naturally produces.

\paragraph{Ecological-style layer.} Material in the style of what an agent natively encounters: document-derived skills rendered from public skill repositories and official skill or API documentation, and background-derived skills distilled from domain background libraries covering the evaluation's languages and cultural regions. Section~\ref{sec:app-audit} shows the genuine material of this kind is scarce for several of our target languages and absent for others; where it is thin, the remaining two layers necessarily carry more of the pool's coverage for that language, approximating what a practitioner assembling a multilingual skill pool today would actually find available.

\paragraph{MT layer.} Produced by translating English source material into multilingual versions, this is the most scalable route to coverage when ecological material is scarce. The layer preserves the cross-lingual transfer artifacts of its production path: \citet{lai2024llms} report that machine-translation quality for low-resource languages lags markedly behind high-resource languages, and \citet{liu2026midb} document that MT content routinely carries content errors, translation defects, and insufficient localization. General Skill Use translates the pool's English sources; Tool Use translates both a schema-derived document and a self-generated document per function into each non-English target; Cultural Skill Use aggregates translated and cross-lingually adapted culture cards according to the contributing resources. Translation uses a single LLM pass (backbone named in Appendix~\ref{sec:app-impl}).

\paragraph{Self-generated layer.} Produced by model self-generation grounded in each domain's task specifications, mirroring the second scalable route commonly used to fill gaps in a scarce multilingual skill ecosystem. \citet{skillsbench2026} find that self-generated skills of this kind provide no average benefit over curated skills and can hurt task success, and \citet{zhang2026coevoskills} note that self-generated material needs verification at construction time; this layer preserves that naturally occurring incompleteness. The generator (backbone named in Appendix~\ref{sec:app-impl}) receives task-type material only: General Skill Use uses source-benchmark task instructions with an explicit instruction to write reusable skill documents; Tool Use uses the function schema plus one example utterance from a held-out split; Cultural Skill Use uses answer-free summaries of its source culture resources. No generator sees an answer key, checker, evaluation utterance, or evaluation outcome, and the Tool Use build additionally scans every pool skill for verbatim evaluation text before accepting the pool.

Every layer passes only a lightweight structural check before inclusion: a well-formed skill body with its target field populated and no empty or truncated content. It is not screened against any test outcome. Table~\ref{tab:arms} summarizes the three layer totals for each domain; the detailed composition follows.

\paragraph{General Skill Use composition.} The 1,750-skill pool contains 400 ecological-style, 550 MT, and 800 self-generated skills. French, Japanese, Korean, and Chinese each receive 100 ecological-style, 75 MT, and 75 self-generated skills, for 250 per language. The ecological-style layer contains no Swahili or Hindi skills, reflecting the community ecosystem audited in Section~\ref{sec:app-audit}; each of these languages instead receives 125 MT and 125 self-generated skills. A further 250 English skills serve as the sources for the translated layer. The six multilingual allocations contribute 1,500 skills, and the English source set brings the pool to 1,750.

\paragraph{Tool Use composition.} The 2,299-skill pool contains 1,254 ecological-style documents, comprising 869 background documents and 385 schema-derived skill documents, together with 660 MT and 385 self-generated skills. The pool draws on the source benchmark's 55-function inventory \citep{kulkarni2025massiveagents}: schema-derived and self-generated material each contributes one document per inventory function in seven languages (385 each), while translating both document types for each function into six non-English targets contributes 660 MT skills. The background-document component supplies the remaining 869 ecological-style entries.

\paragraph{Cultural Skill Use composition.} Cultural Skill Use is organized along culture regions rather than a fixed language roster. Its 5,285-skill pool contains 3,384 ecological-style, 1,283 MT, and 618 self-generated skills. The ecological-style layer aggregates public culture resources and background libraries, the MT layer combines translated and cross-lingually adapted culture cards, and the self-generated layer draws on answer-free summaries; each layer's volume follows the distribution of its contributing resources.

\paragraph{Representative pool entries.} The three excerpts below, one per layer, are drawn from the actual General Skill Use pool. Each is shown as a compact English gloss of its original target-language prose (code and package identifiers are reproduced verbatim; the source language is bracket-tagged).

\begin{quote}\small
\textit{Ecological-style layer, Chinese ([zh]), derived from a public community skill repository:}
\begin{verbatim}
---
name: huashu-prompt-save
description: [zh] Detects a
  prompt's type, saves it.
---
# [zh] Prompt Classification
\end{verbatim}
\end{quote}

\begin{quote}\small
\textit{MT layer, Hindi ([hi]), English source skill translated into Hindi:}
\begin{verbatim}
# [hi] Filling PDF forms
## [hi] Description
[hi] Reads/writes PDF forms
  with interactive fields
  using the `pypdf` library.
pip install pypdf
\end{verbatim}
\end{quote}

\begin{quote}\small
\textit{Self-generated layer, Japanese ([ja]), model self-generation:}
\begin{verbatim}
# SKILL.md: [ja] Induction
## [ja] Overview
[ja] Proves propositions
  about natural numbers in
  Lean 4 via induction
  (`Nat.rec` / `induction`).
\end{verbatim}
\end{quote}

\paragraph{Leakage control.} We separate test-set isolation from legitimate source inclusion in the searchable pool: (1) \textit{Test-set isolation.} For General Skill Use, each query is a new input instance, and we reject any task--skill pair whose skill body contains the exact query, exact answer, or serialized gold object. For Tool Use, evaluation utterances and their gold calls, instance slot values, and checker records are held out from pool construction; skills are built from the function inventory, API and background documents, and source examples outside the evaluation split, followed by a verbatim scan against every evaluation utterance. For Cultural Skill Use, each natural user query is written separately from its source card and audited for source-body 8-gram and answer-key-term overlap; canonical answers, accepted aliases, explicit reject examples, and checker rules remain only in the evaluation manifest. (2) \textit{Skill-pool isolation.} The pool is source-included by design: a reusable skill covering the required procedure, function, or cultural fact may be present because finding and using such material is the object of skill retrieval. The boundary is instance specificity. General and Tool skills may contain reusable procedures, schemas, function names, and slot definitions, but not the evaluation query, serialized gold call, or instance slot values. A Cultural skill may state the fact the task asks the agent to retrieve, but it contains neither the independently written query nor its answer-key and checker metadata. MT and self-generation receive only reusable source material or task-type specifications, never evaluation manifests or downstream outcomes. At inference time, the Theory scorer sees only the anonymous skill body and target locale; the Action scorer sees the user query and anonymous skill body; and the solver sees the task-visible input and selected anonymous skill bodies. Source identifiers, expected-source links, provenance and family labels, gold fields, and checker state remain hidden throughout retrieval, scoring, and solving, and the deterministic checker is applied only after the solver returns its answer.

\section{Implementation Details}
\label{sec:app-impl}

\paragraph{Retrieval.} BM25 applies Unicode word tokenization to each skill's concatenated searchable fields and ranks the raw task query with $k_1=1.5$ and $b=0.75$ \citep{robertson2009bm25}. Neural follows the standard dense-retrieval design \citep{karpukhin2020dpr}, computing dense embeddings over cached \texttt{text-embedding-3-small} vectors and fusing them with lexical and language-match signals in the spirit of sparse-dense hybrid retrieval \citep{luan2021sparse}. SkillFlow follows its released four-stage pipeline \citep{skillflow2025}: five generated search queries retrieve a union with \texttt{bge-base-en-v1.5}, \texttt{ms-marco-MiniLM-L-6-v2} performs shallow reranking, \texttt{bge-reranker-v2-m3} performs deep reranking, and an LLM selector returns final candidates. 

\paragraph{Selection budgets and baselines.} The main comparison reports results at every skill budget $N \in \{1,3,5,10\}$, with Top3 as the primary operating point. Random is averaged over five seeds.

\paragraph{Inference settings.} Theory and Action scoring and the main downstream solver run at temperature at most 0.1 with thinking disabled. For downstream trajectory generalization in Table~\ref{tab:downstream}, we fine-tune Qwen3.5-9B for one epoch with LoRA \citep{hu2022lora} (rank 16, \(\alpha=32\), dropout 0.05, targeting all language-model linear layers). All fine-tuning runs were conducted on a single NVIDIA GeForce RTX 4090 GPU.

\section{Deterministic Checkers}
\label{sec:app-checkers}

All three evaluation domains score task success with a fully deterministic, rule-based checker; no held-out task uses an LLM judge for the headline success/failure outcome. This avoids judge variance and scorer-output self-preference in task success.

\paragraph{General Skill Use.} Each of the 94 tasks carries a short, hand-authored set of gold answer strings: 73 tasks have exactly one gold string, 17 have two, and 4 have three. A response is scored correct if either the full output or any single line of it (splitting on newlines, to tolerate a short preamble or postamble around the actual answer) matches a gold string exactly after normalization: curly quotation marks are converted to straight quotes, internal whitespace runs are collapsed to a single space, and the string is trimmed of surrounding whitespace and one layer of surrounding quote characters. Comparison is otherwise an exact string match, with no case-folding, stemming, or numeric tolerance. Extra gold strings almost always encode benign formatting variance of a single correct answer rather than a semantically different one -- for a spreadsheet lookup-formula task, for instance, the gold set accepts the same \texttt{XLOOKUP} formula with and without a space after each comma, while a formula using the wrong function, cell range, or argument order does not pass.

\paragraph{Tool Use.} The solver must return exactly one well-formed JSON object naming one function and its argument slots, with no markdown fence and no surrounding prose; returning zero or multiple functions, or wrapping the JSON in explanatory text, is rejected outright before any name or slot comparison. The predicted function name must then match the gold name exactly (case-sensitive). Every slot the gold call requires must be present with a matching value, with one exception: a small class of ``defaults to the current moment'' functions may legitimately leave a ``now''-valued slot unstated, mirroring how a real utterance would naturally omit an implicit ``now.'' An extra, unrequested slot is scored as a failure. Slot values (not function names) are normalized before comparison -- Unicode width normalization, lowercasing, whitespace collapsing, and stripping of surrounding punctuation including common CJK punctuation marks -- and a gold slot's accepted value may itself be a short list, so that any one entry counts as correct: a color slot with gold value list \texttt{["red","crimson"]} accepts either name, and a decoratively stylized or full-width prediction normalizes to the same token as a plainly typed one before comparison.

\paragraph{Cultural Skill Use.} Each of the 52 tasks carries a compact, individually authored answer specification: one canonical answer, a short curated list of accepted paraphrases, a short curated list of explicit near-miss wrong answers, and a small backup set of required keywords. A response is checked in three ordered stages after lowercasing, punctuation stripping, and whitespace collapsing. First, if the normalized answer matches (as a substring, in either direction) any explicit wrong-answer example, the task fails immediately, regardless of anything else -- this catches answers that share surface vocabulary with the question but land on the wrong specific fact. Second, if not, the answer is matched against the canonical answer and its paraphrases (again a substring match in either direction); a hit passes. Third, if neither set matches, the checker falls back to counting required keywords, passing only if a minimum count is reached (one keyword suffices when a task defines only one or two; otherwise at least two must appear). For example, a tipping-etiquette question whose canonical answer is a specific percentage range accepts paraphrases restating the same range in different words, while an explicit reject list catches plausible-sounding wrong answers for the same question (e.g.\ ``no tip,'' ``50 percent'') so they cannot pass merely for sharing the word ``tip'' or being a percentage; if the free-form answer matches no listed paraphrase, the keyword backstop requires the numeric range or the word ``tip'' to appear before it can pass.

\section{Limitations and Responsible Use}
\label{sec:app-limitations}

Despite its consistent gains across three evaluation domains and three retrievers, M-SQE has several limitations:

\textbf{Domain Scope.} Our evaluation instantiates three high-frequency skill-use domains, while deployed agents will meet others. The framework is built for this: both scoring views are domain-agnostic --- intrinsic quality and task-grounded fit are properties any skill and any query possess --- and the router's taxonomy reserves the General fusion as the default route for every task that is not tool- or culture-heavy, so an unseen domain is scored by the General path rather than falling outside the method. Specializing further is a five-example change: as Section~\ref{sec:method} notes, a new domain enters the router with five added examples and a fusion adapter, with no retraining of any component.

\textbf{Candidate Coverage.} M-SQE selects within what the retriever surfaces; it does not author new skills, so the pool's supply sets the absolute ceiling. Selection, however, is exactly the lever an agent controls at inference time, and our results show it matters most where supply is thinnest: the languages whose native skill supply the audit measures at zero gain the most from M-SQE, because the usable MT and self-generated candidates that do exist are found rather than lost among relevant-looking alternatives. The downstream trajectory experiment further shows that M-SQE-selected skills make better training material, so quality estimation also strengthens the loop that produces new skills; growing native supply itself remains a task for the broader community, to which we contribute the audit, the pools, and the rubrics.

\textbf{Scorer Generality.} Each reported run computes both views with a single scoring model. The backbone robustness analysis (Fig.~\ref{fig:backbone}) shows the gains survive exchanging the scorer across model families and re-synthesizing the pool, so the finding is not tied to any one scorer; what a single-scorer setup leaves open is aggregation and distillation --- combining scorers, or compressing the released rubrics into a lightweight dedicated model to cut serving cost --- both direct extensions on top of the prompts we release.

\paragraph{Responsible use.} M-SQE targets the language and cultural gaps documented throughout this paper. Our evaluation artifacts (code, skill pools, and tasks) are publicly released at \url{https://github.com/lunyiliu/M-SQE} under an MIT (code) and CC-BY 4.0 (data) license. For culturally sensitive domains we recommend a human or community review pass before deployment; M-SQE's Cultural Skill Use scoring already routes Theory as a factual-reliability check on the retrieved skill (Section~\ref{sec:method}), giving such a review a first-pass filter to build on rather than a blank slate. We further encourage the community to contribute genuinely native-authored skills for the languages our audit finds most scarce.

\section{Case Studies}
\label{sec:app-cases}

The mechanism ablation in Section~\ref{sec:exp} (Table~\ref{tab:ablation}) assigns the two views complementary roles: the Action view carries most of the task fit, while the Theory view guards against the occasional catastrophically flawed candidate. The two cases below further instantiate this role division of two views, with per-task selections and dimension scores taken directly from the evaluation logs; both come from Tool Use under BM25 at Top3.

\paragraph{Case 1: the Action view supplies the missing fit signal.} A task asks the agent to turn on the kitchen lights (gold function \texttt{iot.hue\_lighton}). Theory-only's Top-3 never contains an on-function skill: its rank-1 candidate is a skill for turning off the lights --- which Theory rates a perfect overall quality of 100 (\textit{``clear, uses appropriate terminology\ldots maps perfectly to \texttt{iot.hue\_lightoff}''}). The correct on-function skill, sitting at retrieval rank 7, earns the same 100: Theory judges each skill on its own quality, with no access to the query by design, so sixteen of the twenty candidates tie at 100 and the selection degenerates to retrieval order, whose top ranks are filled by wrong-function lighting skills that lexically overlap the query almost verbatim. With no on-function skill in its Top-3, the solver calls a wrong function and fails. The Action view separates the pair sharply: the off-function skill scores procedure match 0, constraint match 0, and misleading risk 100 (\textit{``the exact opposite of the user's request''}), while the on-function skill scores procedure match 100 and constraint match 100. Action-only and full M-SQE both rank it first and succeed.

\paragraph{Case 2: the Theory view blocks catastrophic content.} A second task asks the agent to post a tweet (gold function \texttt{social.post}). Action-only's Top-3 admits a candidate that names the right function and a slot inventory (\textit{``Function: \texttt{social.post}. Slots: \texttt{business\_name}, \texttt{media\_type}''}) but is otherwise an annotation-style note about the function's applicability, with no executable guidance. The Action view reads it as a near-perfect fit --- procedure match 100, constraint match 100, misleading risk 0, the scorer noting it \textit{``correctly identifies the \texttt{social.post} function and relevant slots''} --- and with the note in context the solver fills the spurious \texttt{business\_name} slot and fails the slot check. Theory reads the same candidate at an overall quality of 10 (\textit{``not a functional tool but a meta-commentary note\ldots useless for execution''}), far below Tool Use's Theory$\geq$65 guard (Section~\ref{sec:method}); full M-SQE therefore excludes it, admits a genuine \texttt{social.post} skill in its place, and succeeds. Fit dimensions read applicability and cannot establish that a document's content will execute; catching exactly that gap is the guard role the Theory view plays.

\end{document}